\documentclass[10pt,twocolumn,letterpaper]{article}

\usepackage[pagenumbers]{cvpr} 

\usepackage{graphicx}
\usepackage{amsmath}
\usepackage{amssymb}
\usepackage{booktabs}

\usepackage{color}
\usepackage{subcaption}
\usepackage{multirow}
\usepackage{wrapfig}
\usepackage{threeparttable}
\usepackage[abs]{overpic}
\usepackage[resetlabels,labeled]{multibib}

\newcites{Supp}{Supplementary}

\usepackage[pagebackref,breaklinks,colorlinks]{hyperref}

\usepackage[capitalize]{cleveref}
\crefname{section}{Sec.}{Secs.}
\Crefname{section}{Section}{Sections}
\Crefname{table}{Table}{Tables}
\crefname{table}{Tab.}{Tabs.}

\def\cvprPaperID{00} 
\def\confName{CVPR}
\def\confYear{2023}

\usepackage[symbol]{footmisc}

\begin{document}

\title{NeuralUDF:
Learning Unsigned Distance Fields\\
for Multi-view Reconstruction of Surfaces with Arbitrary Topologies}

\author{Xiaoxiao Long$^{1,2}$$^{\dagger}$ \quad Cheng Lin$^{2}$$^{\ast}$ \quad Lingjie Liu$^{3}$ \quad Yuan Liu$^{1}$ \quad Peng Wang$^{1}$ \\
Christian Theobalt$^{3}$ \quad Taku Komura$^{1}$ \quad Wenping Wang$^{4}$$^{\ast}$ \\[0.3em]
$^{1}$The University of Hong Kong  \quad $^{2}$Tencent Games \\
$^{3}$Max Planck Institute for Informatics  \quad $^{4}$Texas A\&M University }

\maketitle

\footnotetext[1]{Corresponding authors.}
\footnotetext[2]{This work was conducted during an internship at Tencent Games.}



\begin{abstract}
    We present a novel method, called NeuralUDF, for reconstructing surfaces with arbitrary topologies from 2D images via volume rendering. Recent advances in neural rendering based reconstruction have achieved compelling results. However, these methods are limited to objects with closed surfaces since they adopt Signed Distance Function (SDF) as surface representation which requires the target shape to be divided into inside and outside. In this paper, we propose to represent surfaces as the Unsigned Distance Function (UDF) and develop a new volume rendering scheme to learn the neural UDF representation.
    Specifically, a new density function that correlates the property of UDF with the volume rendering scheme is introduced for robust optimization of the UDF fields.
    Experiments on the DTU and DeepFashion3D datasets show that our method not only enables high-quality reconstruction of non-closed shapes with complex typologies, but also achieves comparable performance to the SDF based methods on the reconstruction of closed surfaces. Visit our project page at \url{https://www.xxlong.site/NeuralUDF/}.
\end{abstract}
\begin{figure}
    \centering
    \begin{overpic}[width=0.95\linewidth]{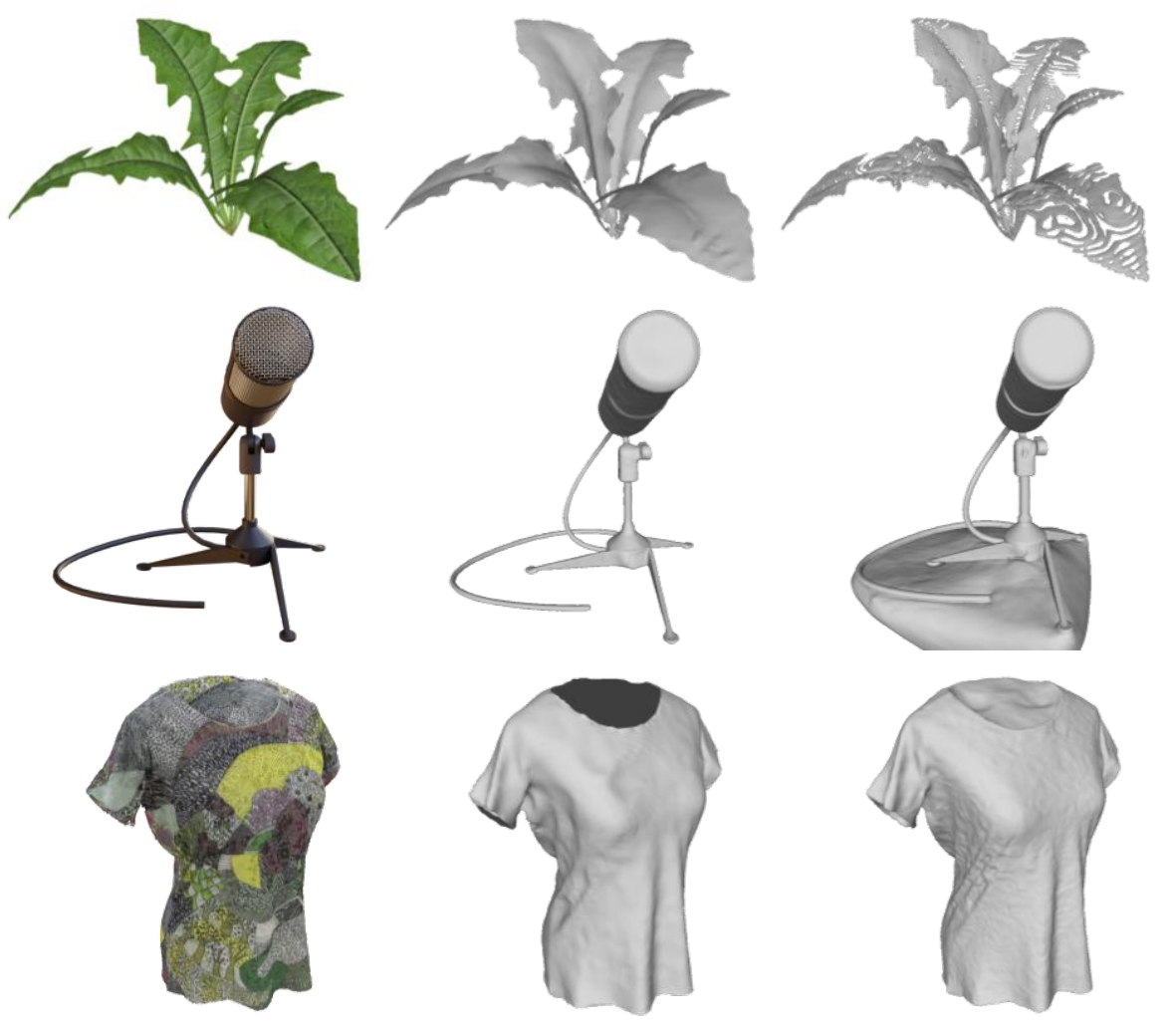}
    \put(10,-5){\small Reference Images}        
    \put(110,-5){\small Ours}
    \put(175,-5){\small NeuS~\cite{wang2021neus}}
    
    \end{overpic}
    \caption{
    We show three groups of multi-view reconstruction results generated by our proposed {\em NeuralUDF} and NeuS~\cite{wang2021neus} respectively.
    Our method is able to faithfully reconstruct the high-quality geometries for both the closed and open surfaces, while NeuS can only model shapes as closed surfaces, thus leading to inconsistent typologies and erroneous geometries. }
    \label{fig:teaser}
    \vspace{-4mm}
\end{figure}
\vspace{-5mm}
\section{Introduction}
Reconstructing high-quality surfaces from multi-view images is a long-standing problem in computer vision and computer graphics. 
Neural implicit fields have become an emerging trend in recent advances due to its superior capability of representing surfaces of complex geometry. NeRF~\cite{mildenhall2020nerf} and its variants~\cite{barron2021mip,lin2021barf,liu2020neural,niemeyer2021giraffe,park2021nerfies,zhang2020nerf++} have recently achieved compelling results in novel view synthesis. For each point in the 3D space, NeRF-based methods learn two neural implicit functions based on volume rendering: a volume density function and a view-dependent color function. Despite their successes in novel view synthesis, NeRF-based methods still struggle to faithfully reconstruct the accurate scene geometry from multi-view inputs, because of the difficulty in extracting high-quality surfaces from the representation of volume density.

VolSDF~\cite{yariv2021volume} and NeuS~\cite{wang2021neus} incorporate the Signed Distance Function (SDF) into volume rendering to facilitate high-quality surface reconstruction in a NeRF framework. However, as a continuous function with clearly defined inside/outside, 
SDF is limited to modeling only closed watertight surfaces.  Although there have been efforts to modify the SDF representation by learning an additional truncation function (e.g., 3PSDF~\cite{chen20223psdf}, TSDF~\cite{sun2021neuralrecon}), their surface representations are still built upon the definition of SDF, thus they are not suitable for representing complex topologies.


We therefore propose to employ the Unsigned Distance Function (UDF) to represent surfaces in  volume rendering. With a surface represented by its zero level set without signs, UDF is a unified representation with higher-degree of freedom for both closed and open surfaces, thus making it possible to reconstruct shapes with arbitrary topologies. 

There are two major challenges in learning a neural UDF field by volume rendering. First, UDF is not occlusion-aware, while the formulation of volume rendering with distance fields requires to estimate surface occlusion for points. Considering a camera ray intersecting with a surface, SDF assumes that the surface is closed and distinguishes the inside/outside with signs, so that a negative SDF value on the ray clearly indicates an occluded point inside the surface. 
In contrast, UDF does not impose the closed surface assumption and always gives non-negative values along the ray. Hence, the UDF value of a point alone cannot be used to infer occlusion. 

The second challenge is that the UDF is not differentiable at its zero-level set. The non-differentiability at the zero-level set imposes barriers in the learning of UDF field. 
The gradients are ill-defined near the iso-surface, leading to difficulty in optimization. 
Consequently, the distance field surrounding the iso-surface is not accurate and the exact zero-level set of a learned UDF cannot be identified in a stable manner.

In this paper, we present a novel method for learning neural UDF fields by volume rendering. We introduce a density function that correlates the property of the UDF representation with the volume rendering process, which effectively tackles the aforementioned challenges induced by the unsigned representation and enables robust learning of surfaces with arbitrary topologies. Experiments on DTU ~\cite{jensen2014large} and DeepFashion3D~\cite{zhu2020deep} datasets show that our method not only enables the high-quality reconstruction of non-closed shapes with complex topologies, but also achieves comparable performance to the SDF based methods on the reconstruction of closed surfaces.

Our contributions can be summarized as:
\begin{itemize}
\vspace{-1mm}
\item We incorporate the UDF into volume rendering, which extends the representation of the underlying geometry of neural radiance fields.
\vspace{-1mm}
\item  We introduce an effective density function that correlates the property of the UDF with the volume rendering process, thus enabling robust optimization of the distance fields.
\vspace{-1mm}
\item Our method achieves SOTA results for reconstructing high-quality surfaces with various topologies (closed or open) using the UDF in volume rendering.

\end{itemize}

\section{Related Work}
\paragraph{Classic multi-view 3D reconstruction.}
3D reconstruction from multiple images is a fundamental problem in computer vision. Traditional reconstruction methods usually adopt discrete representations, such as voxel grids~\cite{ji2017surfacenet,ji2020surfacenet+,kar2017learning,kutulakos2000theory,seitz1999photorealistic,sun2021neuralrecon}, 3D point clouds~\cite{furukawa2009accurate,lhuillier2005quasi}, and depth maps~\cite{campbell2008using,galliani2015massively,schonberger2016pixelwise,tola2012efficient,yao2018mvsnet,yao2019recurrent,gu2020cascade,long2020occlusion,long2021multi,long2021adaptive}. 
The depth map representation is flexible and appropriate for parallel computation, so it is the most widely used representation. 
The depth map based methods first compute depth maps by analyzing the correlation between multiple input images, and then fuse the depth maps into a point cloud, which is then further processed into a mesh via meshing algorithms, such as Poisson surface reconstruction~\cite{kazhdan2006poisson}, ball-pivoting~\cite{bernardini1999ball} and Delaunay triangulation~\cite{lee1980two}.
However, the traditional methods are sensitive to reflections, image noises, and varying illuminations, thus usually yielding incomplete and noisy reconstructions.

\vspace{-4mm}
\paragraph{Neural rendering based surface reconstruction.}
The recent advances on neural implicit representations have achieved promising results in mutli-view 3D reconstruction~\cite{jiang2020sdfdiff,yariv2020multiview,niemeyer2020differentiable,kellnhofer2021neural,liu2020dist,wang2021neus,yariv2021volume,oechsle2021unisurf,zhang2021learning,darmon2021improving}.
These methods typically represent the surfaces to be reconstructed as Signed Distance Function (SDF), and utilize surface rendering or volume rendering to render the pixel colors of the input images for optimizing the SDF fields.
The surface rendering based methods~\cite{niemeyer2020differentiable,yariv2020multiview} require extra masks for optimization and cannot handle objects with complex structures. 
The volume rendering based works~\cite{wang2021neus,yariv2021volume,oechsle2021unisurf,zhang2021learning,darmon2021improving} get rid of the extra masks and achieve more accurate reconstructions.
However, these methods adopt SDF as representation so they are limited to closed surfaces and their performance degrade a lot for non-closed surfaces.


\vspace{-4mm}
\paragraph{Surface representations.} There are various representations to model 3D geometric surfaces. 
NeRF~\cite{mildenhall2020nerf} uses a density field to represent the underlying geometry of a target scene. 
Although NeRF achieves huge successes in novel view synthesis, it's difficult to extract high-quality geometry from such a density representation.
Signed Distance Function (SDF) is widely used for geometry modeling and surface reconstruction~\cite{atzmon2020sal,chen2019learning,gropp2020implicit,mescheder2019occupancy,park2019deepsdf,michalkiewicz2019implicit,wang2021neus,yariv2021volume,oechsle2021unisurf,zhang2021learning,darmon2021improving}, but SDF is only applicable to objects with closed surfaces.
To deal with open surfaces, there are some efforts to learn an additional truncation function (e.g. 3PSDF~\cite{chen20223psdf}, TSDF~\cite{sun2021neuralrecon}). However, their representations are just built on the modification of SDF, and thus are still subject to flexibility of SDF and struggle to model complex surfaces. NDF~\cite{chibane2020neural} uses a neural UDF field to reconstruct surfaces, but they explicitly rely on 3D point clouds for supervising the field optimization.

\section{Method}
Our goal is to reconstruct a surface $\mathcal{S}$ from a set of posed input images $\left\{\mathcal{I}_k\right\}$ of a 3D model. 
We represent the surface as the zero level set of an Unsigned Distance Function (UDF), and propose a novel volume rendering scheme to learn the UDF. We will introduce our method in three parts: 1) We discuss the relationship between the unsigned distance function and the density function of the volume rendering equation;
2) To derive a density function induced by the UDF, we propose a differentiable indicator function to transform the UDF into its associated density field;
3) In addition to the loss terms commonly used in the prior works, we introduce an iso-surface regularizer to improve the stability of the UDF field near the zero level set.
This UDF-based volume rendering scheme enables robust optimization of 
{\em NeuralUDF} for reconstructing both closed surfaces and open surfaces.

\subsection{Volume Rendering}
\noindent
\textbf{Scene representation.} Our method {\em NeuralUDF} represents the scene to be reconstructed by two neural implicit functions: 1) an unsigned distance function $f_u(\mathbf{x})$ that maps a 3D point $\mathbf{x} \in \mathbb{R}^3$ to its unsigned distance to the closest object surface; 2) a view-dependent color function $c(\mathbf{x}, \mathbf{v})$ that encodes the color associated with a point $\mathbf{x} \in \mathbb{R}^3$ and a viewing direction $\mathbf{v} \in \mathbb{R}^3$.

\noindent
\textbf{Volume rendering.} We consider a ray $\mathbf{r}$ emanating from the camera center $\mathbf{o} \in \mathbb{R}^3$ in direction $\mathbf{v} \in \mathbb{R}^3$ ($\|\mathbf{v}\|=1$), which is parameterized by $\{\mathbf{r}(t)=\mathbf{o}+t \mathbf{v} \mid t \geq 0\}$. Volume rendering is used to calculate the color $C$ of the pixel corresponding to the ray $\mathbf{r}$. The accumulated color of the camera ray $\mathbf{r}$ is given by
\begin{equation}
\label{volume_rendering}
C(\mathbf{r})=\int_{0}^{+\infty} T(t) \sigma(t) \mathbf{c}(\mathbf{r}(t), \mathbf{v}) {\rm d}t,
\end{equation}
where $\sigma(t)$ is the volume density, $T(t)=\exp \left(-\int_{0}^t \sigma(u) d u\right)$ denotes the accumulated transmittance along the ray, and the product of $T(t) \sigma(t)$ is the weight function of a point, denoted as $w(t)$, i.e., $w(t) =T(t) \sigma(t)$.

\begin{figure}
    \centering
    \begin{overpic}[width=0.95\linewidth]{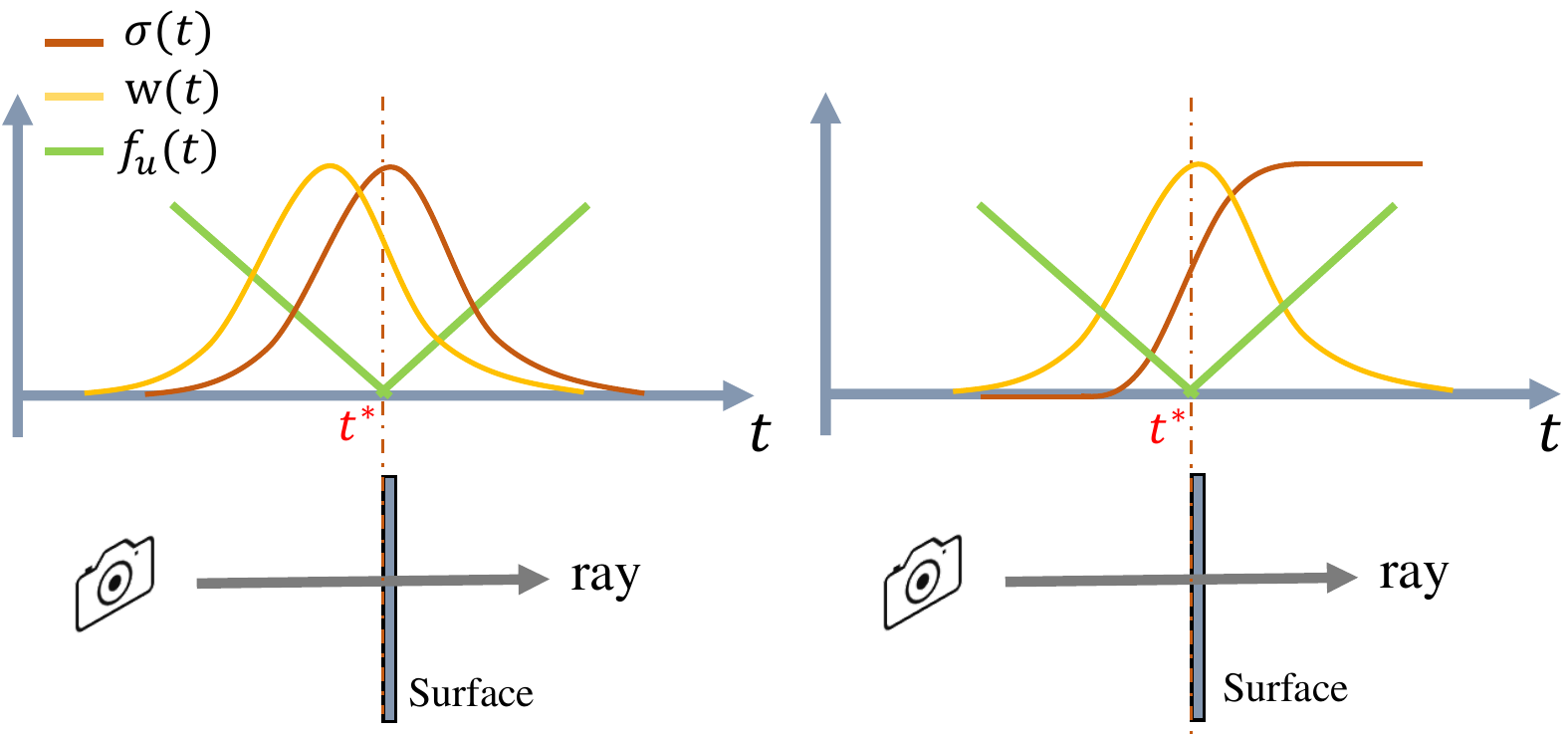}
    \put(0,-5){\small (a) Biased density function}        
    \put(115,-5){\small (b) Unbiased density function}
    
    \end{overpic}
    \caption{
    Illustration of (a) biased density function and (b) unbiased density function for UDF. Suppose that the ray has one single intersection point $t^{*}$ with the surface.
    (a) If the density function is modeled as a bell-shaped probability distribution, the weight function attains a local maximum before $t^{*}$.
    (b) If the density function is modeled as a monotonically increasing function, the weight function can attains local maximum at $t^{*}$.
    }
    \label{fig:illustration}
    \vspace{-4mm}
\end{figure}

\subsection{Modeling Density with UDF.}
In the NeRF framework, the density $\sigma(t)$ is predicted by an MLP, where the points near the surfaces having larger density values and the points far from the surfaces having smaller densities. 
In order to apply volume rendering to learning distance field $f$, we need to define a mapping $\Omega$ to transform the distance function $f(\mathbf{r}(t))$ into a density field $\sigma(t)$, that is $\sigma(t)=\Omega(f(\mathbf{r}(t)))$. 

\vspace{2mm}
\noindent
\textbf{Unbiased volume density function.}
Inspired by NeRF, we may consider defining the density function $\sigma(t)=\Omega(f(\mathbf{r}(t)))$ as a bell-shaped distribution model centered at zero, so that the surface points intersecting with a ray (UDF values are zero) have the maximum density values. 
However, as illustrated in Fig~\ref{fig:illustration}, if the mapping function is straightforwardly defined in this way, a noticeable bias will be induced in the weight function $w(t)$ since its local maximum is not at the intersection point $t^{*}$ where the ray intersects with the surfaces. 
This can result in conspicuous inaccuracy in the reconstructed surfaces, which is demonstrated in~\cite{wang2021neus}.

To define an appropriate mapping, we need to first review the relationship between the density function $\sigma(t)$ and a distance field $f(t)$.
Suppose that there is only one single intersection of the ray and a surface. It is shown in NeuS~\cite{wang2021neus} that, to make the weight function $w(t)$ unbiased, the density function should be in the form of a \textbf{\em monotonically increasing function}, which attains the maximum at an infinite point rather than at the intersection point.


Accordingly, NeuS~\cite{wang2021neus} chooses to use a Sigmoid function $\Phi_{\kappa}(x)=\left(1+e^{-\kappa x}\right)^{-1}$ to achieve the goal. 
The accumulated transmittance $ T(t)$ is then defined as $ T(t)=\Phi_{\kappa}(f_s(t))$, where $f_s(t)$ is the signed distance function (SDF). 
The density function $\sigma_s(t)$ induced by SDF is formulated as
\begin{equation}
\sigma_s(t)=\Omega_s(f_s(\mathbf{r}(t)))
    =\kappa|\cos(\theta)| \left(1-\Phi_{\kappa}(f_s(\mathbf{r}(t)))\right),
\label{eq:sdf_density}
\end{equation}
where $\kappa$ is the parameter of Sigmoid function $\Phi_{\kappa}(x)$, and $\theta$ is the angle between the view direction $\textbf{v}$ and the surface normal $\mathbf{n}$. 

\vspace{2mm}
\noindent
\textbf{Unbiased volume density for UDF.} However, the unbiased density formulation induced by SDF~\cite{wang2021neus} cannot be directly applied to UDF.
The formulation of UDF near the surface can be given by
\begin{equation}
f_{u}(\mathbf{r}(t))=\left\{\begin{array}{ll}
|\cos (\theta) \cdot (t-t^{*})|, & t \neq t^{*} \\
0, & t = t^{*}
\end{array}\right.
\end{equation}
where $\theta$ is the angle between the view direction $\textbf{v}$ and the surface normal $\mathbf{n}$. 
Obviously, the UDF value is not monotonous along a ray when it crosses a surface, where the mapping $t \rightarrow f_u(\mathbf{r}(t)) $ is non-injective. This leads to difficulties in modeling the monotonicity between $t$ and the density function $\sigma(t) = \Omega(f_u(\mathbf{r}(t)))$, given that an unbiased density function is expected to be monotonically increasing.

To leverage the unbiased property derived from the monotonic density function, we need to model the injective mapping from $t$ to the density function $\sigma_u(t)$ with the formulation of UDF $f_u(\mathbf{r}(t))$. Since the monotonicity of UDF along a ray only changes at the intersection point $t^*$, we can use a visibility indicator function $\Psi(t)$ to reflect the existence of the intersection and thus capture the monotonicity change:

\begin{equation}
\label{indicator_hint}
\Psi(t)=\left\{\begin{array}{ll}
1, & t < t^{*} \\
0, & t > t^{*}.
\end{array}\right.
\end{equation}
The indicator function $\Psi(t)$ indicates the visibility of the points sampled along the ray $\mathbf{r}$, where 1 means that the point is visible in the free space while 0 means that the point is occluded. Note this just gives a simple hint for the indicator function, while the actual formulation will be derived later.

With $\Psi(t)$, given an unbiased density function $\sigma(t) = \Omega_s(f(\mathbf{r}(t)))$ that is monotonically increasing, the density function for UDF can be defined as 

\begin{equation}
\label{udf_density_func}
\begin{split}
\sigma_u(t)&=
\Psi(t) \Omega_s(f_u(\mathbf{r}(t))) 
+ (1-\Psi(t)) \Omega_s(-f_u(\mathbf{r}(t))) , \\
\end{split}
\end{equation}
where $\Omega_s$ can be any monotonic density function that is unbiasedly designed, while in this paper we directly use the formulation of Eq.\ref{eq:sdf_density}. Since unsigned distance function $f_u$ is not differentiable at $t^{*}$, the domain scope of $\sigma_u(t)$ is defined as  $ \{t \in \mathbb{R}^{1}, t\neq t^{*}\}$.


\vspace{2mm}
\noindent
\textbf{Zero level set.} The visibility indicator function $\Psi$ plays a vital role in the formulation of volume density for UDF, because it essentially encodes the distribution of the geometric surface. The key to the visibility is to find the intersection point  $t^{*}$ of surfaces and the ray, whose UDF value is zero, i.e., $f_{u}(\mathbf{r}(t^{*}))=0$. 

It is nontrival to find the zero level set for a UDF field.
In practice, it's impossible to locate a point with exact zero UDF value, and a trade-off to balance accuracy and efficiency is to find points whose UDF values are smaller than a pre-defined threshold $\epsilon$.
As a result, the approximated surfaces will be relaxed from exact zero level set to be a $\epsilon$-bounded surface band $\mathcal{S}=\left\{\mathbf{x} \in \mathbb{R}^3 \mid f_{u}(\mathbf{x})< \epsilon \right\}$.

However, such a definition of the indicator function involving $\epsilon$ bounding brings about the following difficulties in learning the UDF field.
This indicator function design with $\epsilon$ bounding operation incurs a set of barriers that impede learning the UDF field:
1) choosing an appropriate $\epsilon$ is difficult; 
2) a fixed $\epsilon$ cannot be adjusted to adapt the UDF field in different optimization stages;
3) the step-like indicator function is not differentiable. These limitations make the optimization of UDF unstable and easy to collapse. 

\vspace{2mm}
\noindent
\textbf{Visibility indicator function.}
To overcome the limitations mentioned above, we leverage probabilistic models to make the indicator function $\Psi(t)$ differentiable.
Therefore, we propose to approximate the surfaces by a soft probabilistic distribution instead of using a hard $\epsilon$ bound. We measure the probability of a point $\mathbf{r}(t)$ being on a surface, named as surface existence probability $h(\mathbf{r}(t))$ abbreviated as $h(t)$. 
This can be achieved by transforming the UDF values into probabilities via a bell-shaped distribution centered at 0. Here we use a logistic distribution $\phi_{\beta}(x)$ to define the probability $h(t_{i})$:
\begin{equation}
\label{existence_probability}
    h(t_{i})=1-\exp \left(-\alpha \cdot\phi_{\beta}(f_u(\mathbf{r}(t_i))) \cdot \delta_i\right),
\end{equation}
\begin{equation}
\label{logistic_density_func}
    \phi_{\beta}(x)= \beta e^{-\beta x} /\left(1+e^{-\beta x}\right)^2
\end{equation}
where $t_i$ is the $i$-th sample in the ray $\mathbf{r}$, $h(t_{i})$ denotes the surface existence probability of the sample $t_i$,
$\delta_i=t_{i+1}-t_i$ is the distance between adjacent samples, and $\phi_{\beta}(t)$ is a logistic distribution function.
Note that the standard deviation of $\phi_{\beta}(x)$ is given by $1/\beta$, where $\beta$ is a trainable parameter, and $1/\beta$ approaches to zero as the network training converges; $\alpha$ is a constant scalar to scale the distribution.

After obtaining the surface existence probabilities of the sampled points on the ray $\mathbf{r}$, we can measure the probability of whether a point is occluded by any surfaces or not. Since local UDF values do not indicate occlusion, we accumulate the probabilities of the points $\mathbf{r}(t_j)$ ahead of a query point $\mathbf{r}(t_i)$ along the ray:

\begin{equation}
\label{soft_ind_0}
\hat{\Psi}(t_i)=\prod_{j=1}^{i-1}\left(1-h(t_j) \right),
\end{equation}
where $h(t_i) \in [0, 1]$ measures the probability of a point being on the surfaces.
From this equation, we can see that $\hat{\Psi}(t_i)$ will drop dramatically to zero once there is a point that has high surface existence probability on the ray. Compared to using a hard bound, the probability accumulation strategy makes the indicator function $\hat{\Psi}(t)$ differentiable and learnable, which adapts to learning various surface geometries. The function $\hat{\Psi}(t)$ will become sharper when $1 / \beta$ approaches 0 as the network training converges, resulting in more accurate approximation of the zero level set.

\begin{figure}[tp!]
    \centering
    \begin{overpic}[width=\linewidth]{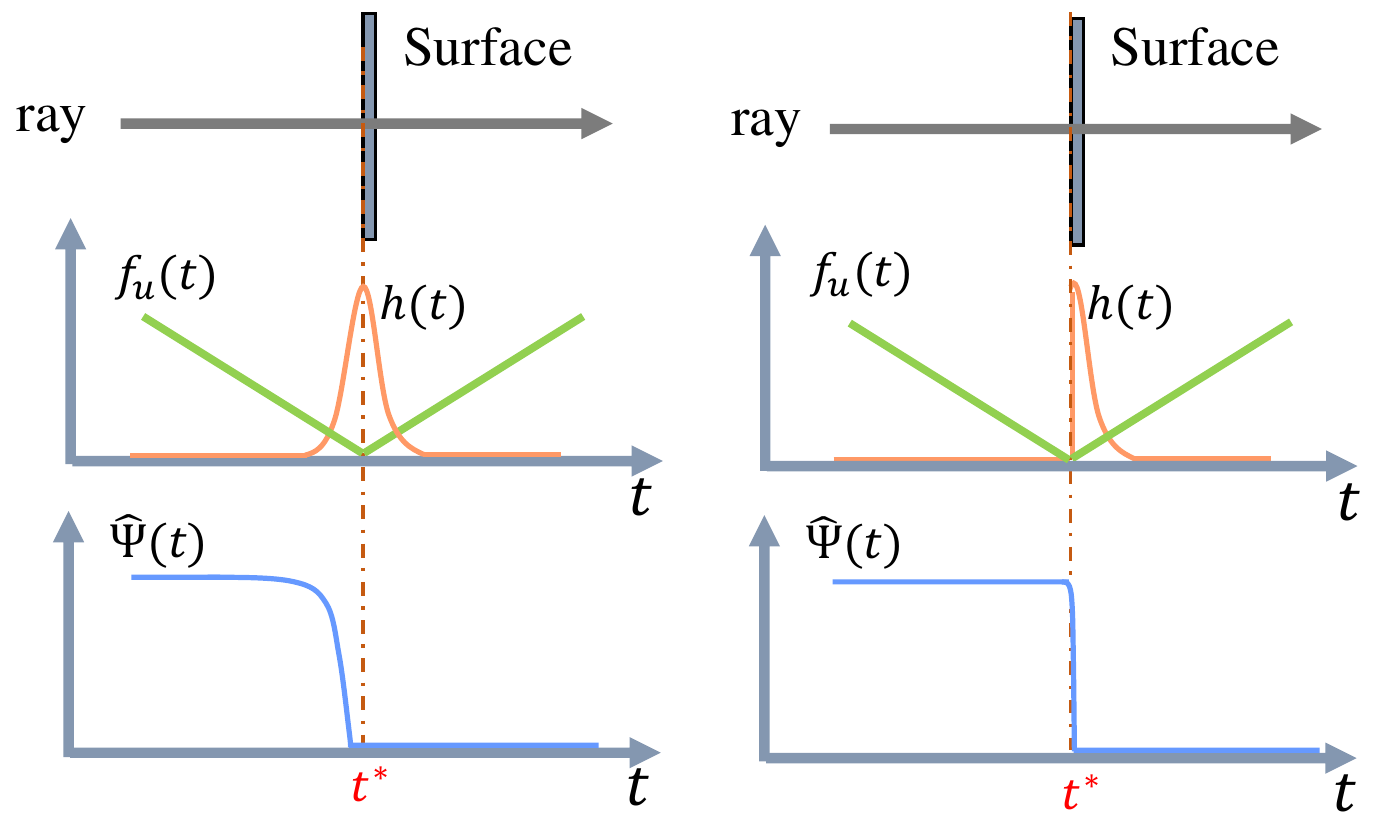}
    \put(15,-6){\small (a) Naive indicator func.}    
    \put(125,-6){\small (b) Grad-aware indicator func.}
    \end{overpic}
    
    \caption{
    Illustration of (a) naive visibility indicator function and (b) gradient-aware visibility indicator function.
    The naive indicator function will attain zero before the ray intersects with surface point $t^{*}$. Incorporating the UDF gradients information, the gradient-aware indicator function attains zero exactly at the intersection point $t^{*}$.
    }
    \label{fig:indicator_func}
    \vspace{-4mm}
\end{figure}

\vspace{2mm}
\noindent
\textbf{Gradient-aware indicator.} 
Although the formulation of $\hat{\Psi}(t)$ in Eq.~\ref{soft_ind_0} is effective for identifying surfaces in learning the UDF field, it can induce bias in the indicator.
As shown in Figure~\ref{fig:indicator_func}, the surface existence probability $h(t_i)$ increases when the ray approaches a surface, and the accumulated probability $\hat{\Psi}(t_i)$ can attain zero before the point $t^{*}$ where the ray intersects with the surface. 

To alleviate this problem, we modify the indicator function $\hat{\Psi}(t)$ with a masking function $m(t)$
\begin{align}
    \label{soft_ind}
\hat{\Psi}(t_i)=\prod_{j=1}^{i-1}\left(1-h(t_j) \cdot m(t_j) \right) \\
m(t_j)=\left\{\begin{array}{ll}
0, & \cos(\theta_{j+1}) < 0 \\
1, & \cos(\theta_{j+1}) \geq 0
\end{array}\right.
\end{align}
where $\theta_{j+1}$ denotes the angle between the ray direction and the gradient of $f_u(\mathbf{r}(t_{j+1}))$. This formulation exploits the geometric property of UDF surface, that is, the gradient direction of UDF flips across its zero level set. Therefore, the points with negative $\cos (\theta_{j+1})$ are located ahead of the intersection point $t^{*}$; we mask out these points and only accumulate the surface existence probability once the ray reaches the surface. The final formulation jointly leverages the insights from both UDF value and gradient, leading to more reliable surface identification.

\begin{figure*}[tp!]
    \centering
    \begin{overpic}[width=\linewidth]{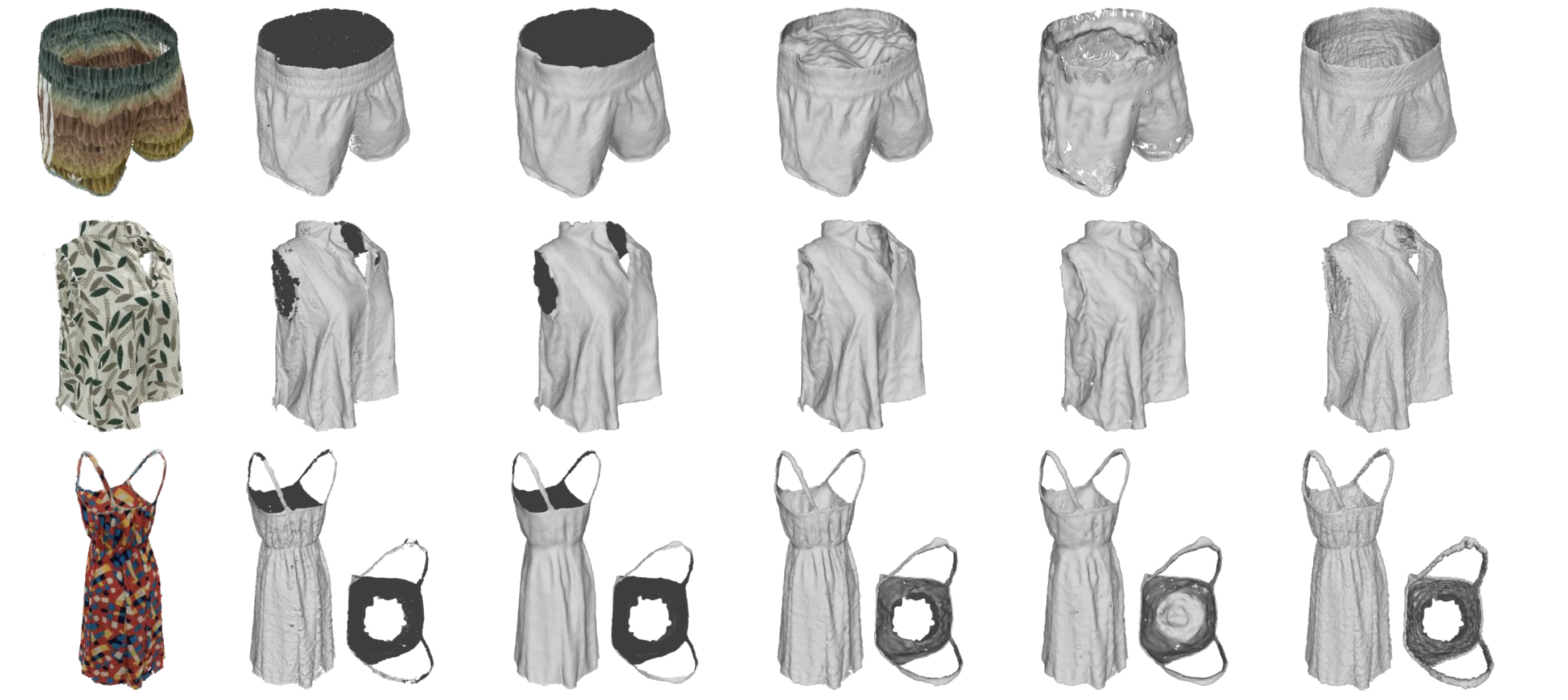}
    \put(5,-5){\small Reference Images}  
    \put(85,-5){\small GT meshes}
    \put(180,-5){\small Ours}
    \put(255,-5){\small NeuS~\cite{wang2021neus}}
    \put(335,-5){\small VolSDF~\cite{yariv2021volume}}
    \put(425,-5){\small NeRF~\cite{mildenhall2020nerf}}
    \end{overpic}
    \caption{
    Qualitative comparisons with SOTAs on DeepFashion3D~\cite{zhu2020deep} dataset.
    The GT meshes are reconstructed from the GT point clouds provided by DeepFashion3D dataset via Ball Pivoting meshing algorithm~\cite{bernardini1999ball}.
    Thanks to the UDF representation, {\em NeuralUDF} accurately recovers the garments with open surfaces. However the SDF-based methods can only model the garments as closed surfaces, thus leading to inconsistent topologies and erroneous geometries. (See supplementary materials for more results.)
    }
    \label{fig:garment_compare}
\end{figure*}
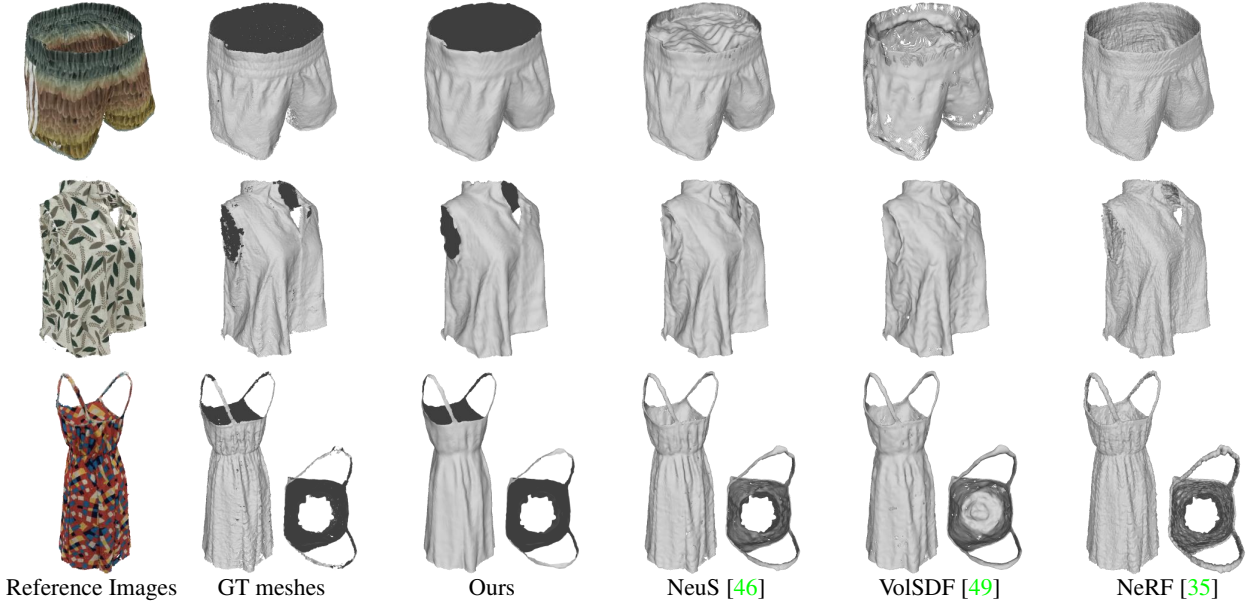
\begin{table*}[ht!]
\begin{center}
\setlength{\abovecaptionskip}{4pt}
\resizebox{\linewidth}{!}{%
\begin{threeparttable}[b]
\begin{tabular}{l|ccccc ccccc ccccc | c}
\hline
Scan & LS-C0 & SS-D0 & NS-D0 & NS-C0 & Skirt0 & LS-D0 & Trouser0 & NS-D1 & Trouser1 & LS-C1 & Skirt1 & SS-C0  & Mean \\
\hline
Colmap~\cite{schonberger2016structure} & 2.95 & 2.91 & 3.58 & 3.06 & 3.23 & \textbf{3.26} & 3.09 & 3.11 & 2.95 & 3.16 & 2.97 & 2.95 & 3.10\\
IDR\tnote{$\dagger$}~\cite{yariv2020multiview} & 10.85 & 8.20 &  5.00 &  9.63 & 6.76 & 6.33 & 6.32 & 3.32 & 3.75 & 6.08 & 2.44 & 7.02 & 6.31 \\
\hline
Nerf~\cite{mildenhall2020nerf} & 6.38 & 7.84 & 5.03 & 5.45 & 4.69 & 7.86 & 4.82 & 4.63 & 6.04 & 7.17 & 8.24 & 6.89 & 6.25 \\
VolSdf~\cite{yariv2021volume} & 5.92 & 4.79 & 4.54 & 8.57 & 7.02 & 5.96 & 6.97 & 4.36 & 7.24 & 8.73 & 7.74 & 8.84 & 6.72 \\
Neus~\cite{wang2021neus} &  3.18 & 4.82 & 4.78 & 4.99 & 3.73 & 5.71 & 5.89 & 2.21 & 5.89 & 3.60 & 2.44 & 5.13 & 4.36\\
NeuralWarp~\cite{darmon2022improving} & 4.71 & 11.24 & 5.40 & 8.56 & 4.10 & 5.83 & 5.79 & 2.00 & 2.07 & 6.50 & 9.61 & 6.94 & 6.06\\
Ours & \textbf{1.92 }& \textbf{2.05} & \textbf{2.36} & \textbf{1.58} & \textbf{1.33} & \underline{4.11} & \textbf{2.47} & \textbf{1.50} & \textbf{1.63} & \textbf{2.47} & \textbf{2.16} & \textbf{2.15} & \textbf{1.97}\\
\hline
\end{tabular}%

\begin{tablenotes}
     \item[$\dagger$] Optimization using extra object masks.
     \item LS-C denotes \textbf{L}ong-\textbf{S}leeve \textbf{C}oat, SS-C denotes \textbf{S}hort-\textbf{S}leeve \textbf{C}oat, 
     NS-C denotes \textbf{N}one-\textbf{S}leeve \textbf{C}oat, 
    LS-D denotes \textbf{L}ong-\textbf{S}leeve \textbf{D}ress, SS-D denotes \textbf{S}hort-\textbf{S}leeve \textbf{D}ress, 
         NS-D denotes \textbf{N}one-\textbf{S}leeve \textbf{D}ress. 
   \end{tablenotes}
\end{threeparttable}

}
\caption{Quantitative evaluation results on Deepfashion3D~\cite{zhu2020deep} dataset.}
\label{tab: deepfashion_eval}
\end{center}
\vspace{-6mm}
\end{table*}

\subsection{Optimizing UDF Fields.}
\paragraph{Iso-surface regularizer.}
With the volume rendering scheme mentioned above, {\em NeuralUDF} learns a UDF field from the images by minimizing the differences between rendered images and ground truth images. 
However, unlike SDF representation that is defined for closed surface, the intrinsic property of the UDF field makes it more difficult to optimize. First, the UDF has a higher degree of freedom accommodating arbitrary topologies instead of only closed surfaces.
Second, the UDF is not differentiable at the zero level set; the values and gradients of the UDF field close to the zero level set are therefore hard to learn. 
This usually results in a wide strip around the iso-surface where the UDF values are distributed irregularly. Therefore, we introduce a simple yet effective geometric regularizer term during training to improve the stability of the UDF field:
\begin{equation}
\label{iso_regularizer}
\mathcal{L}_{\text {reg}}=\frac{1}{M} \sum_{k,i} \exp (-\tau \cdot f_u(\mathbf{r}_{k}(t_{k,i}))),
\end{equation}
where $\tau$ is a constant scalar to scale the UDF values, and $M$ is the number of sampled rays per optimization step. This formulation effectively precludes the UDF values of the non-surface points from being zero, which thus encourages the simplicity and clearness of the zero level set in the field. The experiments show that this iso-surface regularizer excels at improving the surface quality (see Sec.~\ref{ablation}).


\vspace{2mm}
\noindent
\textbf{Loss functions.}
We train {\em NeuralUDF} by enforcing the consistency of the rendered colors and the ground truth colors of the input images without using 3D ground truth shapes. 
Specifically, we optimize our neural networks and two trainable probabilistic parameters $\kappa$ of Eq.~\ref{eq:sdf_density} and $\beta$ of Eq.~\ref{logistic_density_func} by randomly sampling a batch of pixels and their corresponding rays.

The overall loss  function is defined as a weighted sum of several loss terms:
\begin{equation}
    \mathcal{L}=\mathcal{L}_{\text {color }} + \lambda_{0} \mathcal{L}_{\text {patch }} +\lambda_{1} \mathcal{L}_{\text {eik }} +\lambda_{2} \mathcal{L}_{\text {reg}} \left(+  \gamma \mathcal{L}_{\text {mask }} \right).
\end{equation}
where $\mathcal{L}_{\text{color}}$ is the L1 color loss to minimize the differences between the rendered pixel colors and the ground truth pixel colors of the input images.
Following the prior works~\cite{yariv2020multiview,wang2021neus,yariv2021volume}, the Eikonal term $\mathcal{L}_{eik}$ is used to regularize the neural distance field to have unit norm of gradients, and the optional mask loss term $\mathcal{L}_{\text {mask}}$ can be included if extra masks are provided.
$\mathcal{L}_{\text {patch}}$ is a patch-based color loss term introduced in~\cite{long2022sparseneus}, which renders the colors of a local patch of an input image using patch blending strategy and then enforces the consistency of the rendered colors and ground truth colors at the patch level. (See the supplementary materials for the details.)

\section{Experiments}

\begin{figure*}
    \centering
    \begin{overpic}[width=0.95\linewidth]{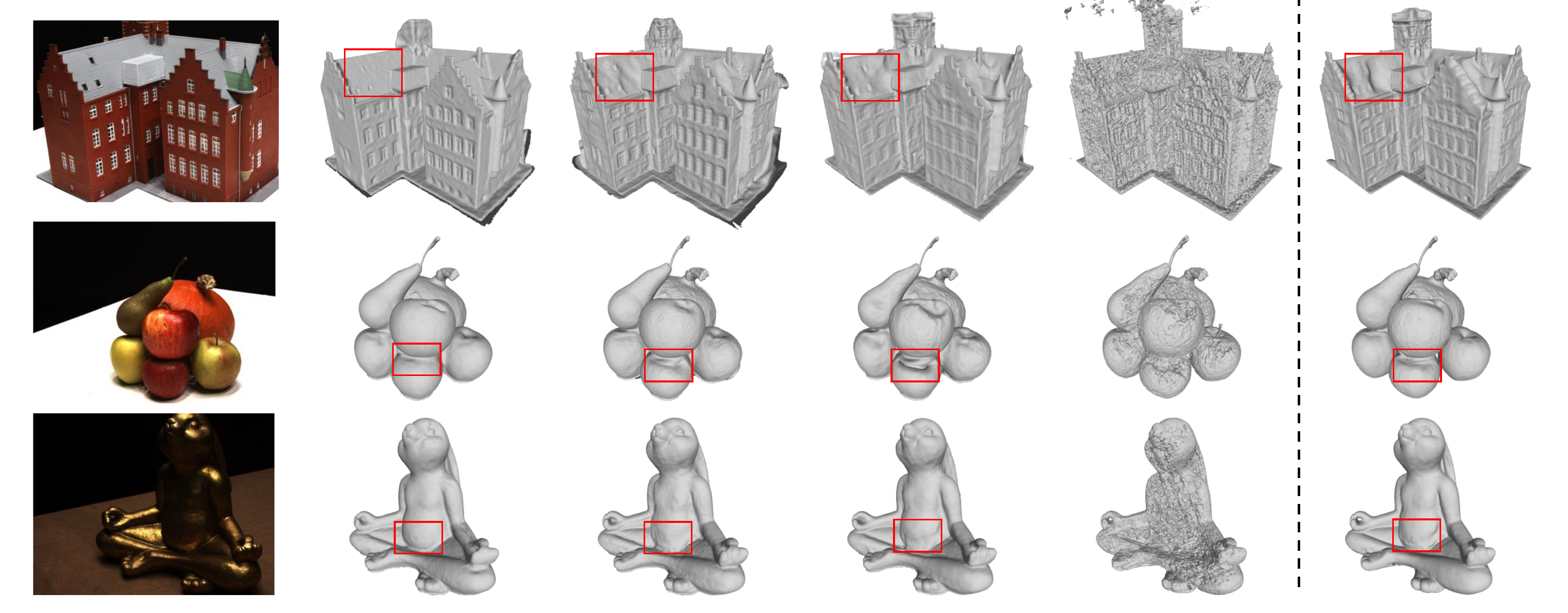}
    \put(15,-4){\small Reference Images}  
    \put(120,-4){\small Ours}
    \put(185,-4){\small NeuS~\cite{wang2021neus}}
    \put(257,-4){\small VolSDF~\cite{yariv2021volume}}
    \put(335,-4){\small NeRF~\cite{mildenhall2020nerf}}
    \put(412,-4){\small IDR~\cite{yariv2020multiview}}
    \end{overpic}
    \caption{
    Qualitative comparisons with SOTAs on DTU~\cite{jensen2014large} dataset.(See the supplementary materials for more results.)
    }
    \label{fig:dtu_compare}
\end{figure*}
\begin{table*}[t]
\setlength{\abovecaptionskip}{1pt}
\begin{center}
\resizebox{0.95\linewidth}{!}{%
\begin{threeparttable}[b]
\begin{tabular}{l|ccccc ccccc ccccc | c}
\hline
Scan &24 & 37 & 40 & 55 & 63 & 65 & 69 & 83 & 97 & 105 & 106 & 110 & 114 & 118 & 122 & Mean \\
\hline
Colmap~\cite{schonberger2016structure} & 0.81  &2.05  & 0.73  & 1.22 & 1.79  & 1.58  & 1.02  & 3.05  & 1.40  & 2.05  & 1.00 & 1.32  & 0.49 & 0.78 & 1.17 & 1.36 \\
IDR\tnote{$\dagger$}~\cite{yariv2020multiview} &  1.63 & 1.87 & 0.63 & 0.48 & 1.04 & 0.79 & 0.77 & 1.33 & 1.16 & 0.76  & 0.67 & 0.90 & 0.42 & 0.51 & 0.53 & 0.90\\
\hline
Nerf~\cite{mildenhall2020nerf} & 1.90 & 1.60 & 1.85 & 0.58 & 2.28 & 1.27 & 1.47 & 1.67 & 2.05 & 1.07 & 0.88 & 2.53 & 1.06 & 1.15 & 0.96 & 1.49 \\
UniSurf~\cite{oechsle2021unisurf} & 1.32 & 1.36 & 1.72 & 0.44 & 1.35 & 0.79 & 0.80 & 1.49 & 1.37 & 0.89 & \underline{0.59} & 1.47 & 0.46 & 0.59 & 0.62 & 1.02 \\
VolSdf~\cite{yariv2021volume} & 1.14 & 1.26 & 0.81 & 0.49 & 1.25 & \underline{0.70} & \underline{0.72} & 1.29 & 1.18 & \underline{0.70} & 0.66 & 1.08 & 0.42 & 0.61 & \underline{0.55} & 0.86 \\
Neus~\cite{wang2021neus} & 1.37 & 1.21 & 0.73 & \underline{0.40} & 1.20 & \underline{0.70} & \underline{0.72} & \textbf{1.01} & 1.16 & 0.82 & 0.66 & 1.69 & \underline{0.39} & \textbf{0.49} & \textbf{0.51} & 0.87  \\
NeuralWarp~\cite{darmon2022improving} & \textbf{0.49} & \textbf{0.71} & \textbf{0.38} & \textbf{0.38} & \textbf{0.79} & 0.81 & 0.82 & \underline{1.20} & \underline{1.06} & \textbf{0.68} & 0.66 & \textbf{0.74} & 0.41 & 0.63 & \textbf{0.51} & \textbf{0.68}  \\

Ours & \underline{0.69} & \underline{1.18} &	\underline{0.67} &	0.44 &	\underline{0.90} &	\textbf{0.66} &	\textbf{0.67} &	1.32 &	\textbf{0.94} &	0.95 &	\textbf{0.57} &	\underline{0.86} &	\textbf{0.37} &	\underline{0.56} &	\underline{0.55} &	\underline{0.75} \\
\hline
\end{tabular}%

\begin{tablenotes}
     \item[$\dagger$] Optimization using extra object masks.
   \end{tablenotes}
\end{threeparttable}
}
\caption{Quantitative evaluation results on DTU~\cite{jensen2014large} dataset.}
\label{tab: dtu_eval}
\end{center}
\vspace{-8mm}
\end{table*}
\subsection{Experimental settings}
\noindent
\textbf{Datasets.}
We conduct evaluations on the DTU~\cite{jensen2014large} dataset and Deepfashion3D~\cite{zhu2020deep} dataset.
DTU is a multi-view dataset containing objects with closed surfaces, and each scene consists of 49 or 64 images with $1600 \times 1200$ resolution. We use the same 15 scenes following the prior works for comparison.
To validate the effectiveness of our method on objects with open surfaces, we adopt Deepfashion3D dataset for evaluation.
Deepfashion3D is a large-scale real 3D garment dataset containing 563 diverse cloth items, and provides the ground truth point clouds of the captured garments. 
To facilitate the multi-view 3D reconstruction task, we randomly choose 12 garments from different categories of the dataset, and then scale the point clouds to be in the unit sphere.
We render each point cloud into 72 images with $1024 \times 1024$ resolution. The rendered images are used for reconstruction.
We further collect a set of synthetic plant objects for further qualitative comparisons. Each plant object is rendered into 324 images with $256 \times 256$ resolution. 

\vspace{2mm}
\noindent
\textbf{Implementation details.}
We use MLPs to model the unsigned distance field and color radiance field.
We use the Adam optimizer with a global learning rate 5e-4 for the network training, specially we set the learning rate of the UDF MLP as 1e-4. We sample 512 rays per batch and train our model for 300k iterations on a single NVIDIA RTX 2080Ti GPU. Our network architecture and initialization scheme are similar to those of prior works~\cite{yariv2020multiview,wang2021neus,yariv2021volume}.
We leverage MeshUDF~\cite{guillard2021meshudf}, an extended marching cube algorithm designed for UDF, to extract explicit mesh from the learned UDF field.

A hierarchical sampling strategy is used to sample points along a ray in a coarse-to-fine manner for volume rendering. We first uniformly sample points on the ray and then iteratively conduct importance sampling~\cite{mildenhall2020nerf} to sample more points on top of coarse probability estimation. The probability of coarse sampling is computed based on the density function $\sigma_{u}(t)$ with fixed and predefined parameters $\kappa$ and $\beta$, while the probability of fine sampling is computed with the learned $\kappa$ and $\beta$.

\noindent
\textbf{Baselines.}
We compare {\em NeuralUDF} with the following state-of-the-art baselines: 1) a widely used classic reconstruction method COLAMP~\cite{schonberger2016structure}, which we use to reconstruct a mesh from the point cloud generated by COLMAP via Screened Poisson Surface Reconstruction~\cite{kazhdan2013screened}; 2) neural implicit methods utilizing surface rendering, IDR~\cite{yariv2020multiview} and UNISURF~\cite{oechsle2021unisurf}; 3) neural implicit methods utilizing volume rendering, NeuS~\cite{wang2021neus}, VolSDF~\cite{yariv2021volume} and NeuralWarp~\cite{darmon2022improving}; 4) NeRF~\cite{mildenhall2020nerf}.
Although NeRF is not designed for surface reconstruction, we include NeRF for comparisons since its density representation enables successful novel view synthesis on objects with arbitrary topologies. 
Following NeuS~\cite{wang2021neus}, we use a density threshold of 25 on DTU (50 on Deepfashion3D) to extract surfaces from NeRF for comparisons.
Note that, IDR takes extra masks as input for optimization while ours and the other methods do not.

\subsection{Comparisons}

\noindent
\textbf{Evaluation on non-closed objects.}
We conduct evaluations on DeepFashion3D dataset whose objects are non-closed, and measure the Chamfer Distances ($10^{-3}$) between the predicted meshes with ground truth point clouds. 
As shown in Table~\ref{tab: deepfashion_eval}, our method outperforms the other methods by a large margin. 
This is because our method can accurately reconstruct the surfaces with arbitrary topologies while the SDF based methods can only model the garments as closed surfaces and thus exhibit large errors. Note although IDR takes extra masks for optimization, the reconstructed results still exhibit large quantitative errors.

In Figure~\ref{fig:garment_compare}, the SDF based methods including NeuS, VolSDF and IDR enforce strong assumption on the closeness of the target shape and attempt to model all the shapes as closed surfaces, thus inevitably leading to large errors and inconsistent topologies for the garments. For example, for the first row, VolSDF attempts to use a double-layered surface to represent the thin cloth; optimizing this tiny SDF band is difficult and the results contain many noticeable holes. For the second row, NeuS gives a closed surface for the garment model while the cuffs are even incorrectly connected. 


\vspace{1mm}
\noindent
\textbf{Evaluation on closed objects.}
We further perform evaluation on DTU dataset whose objects are closed and watertight, and measure the Chamfer Distances between the predicted meshes and ground truth point clouds.
Although our method adopts UDF as representation and does not impose closed surface assumption, our method still achieves comparable performance with the SDF based methods that are particularly designed for closed shapes.

As shown in Table~\ref{tab: dtu_eval}, in terms of the averaged Chamfer Distance, {\em NeuralUDF} outperforms most of the SOTA methods except the most recent work NeuralWarp~\cite{darmon2021improving}. 
NeuralWarp is a fine-tuning method that further optimizes the surfaces obtained from VolSDF~\cite{yariv2021volume} by minimizing the differences between the warped source images and the reference image.
However, NeuralWarp heavily relies on the reconstructed geometry provided by other methods to generate reasonable warped source images, so it cannot be trained from scratch. 
The qualitative comparisons with SOTAs are provided in Figure~\ref{fig:dtu_compare},
where our reconstruction results contain fewer artifacts than the SDF based methods, e.g., the roof of the House and the Apple.

\subsection{Ablations and Analysis}
\label{ablation}

\begin{figure}[tp!]
    \setlength{\abovecaptionskip}{20pt}
    \centering
    \begin{overpic}[width=\linewidth]{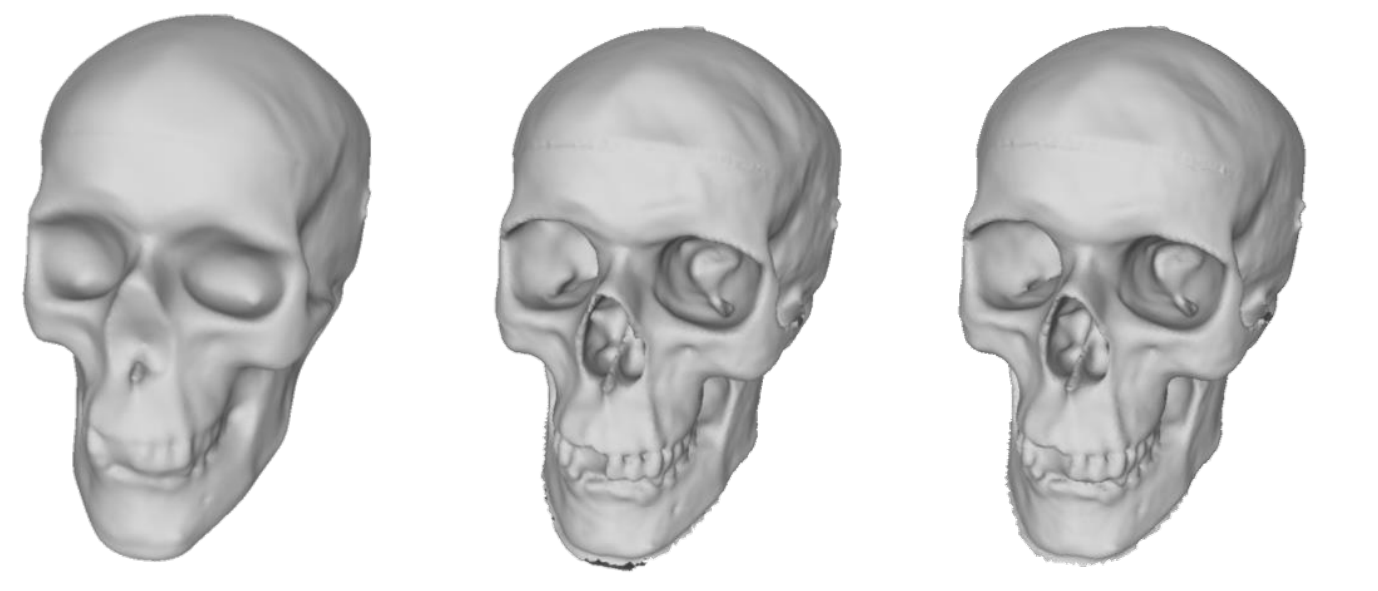}
    \put(40,7){CD: 1.31}        
    \put(122,7){CD: 0.75}
    \put(202,7){CD: 0.66}
    
    \put(0,-5){\footnotesize (a) Naive indicator}      
    \put(12,-13){\footnotesize function}     
    \put(82,-5){\footnotesize (b) Gradient-aware}
    \put(93,-13){\footnotesize indicator function}
    \put(165,-5){\footnotesize (c) Patch blending}

    \end{overpic}
    \caption{
    Ablation studies on gradient-aware visibility indicator function and patch blending loss term.
    }
    \label{fig:ablation}
    \vspace{-4mm}
\end{figure}
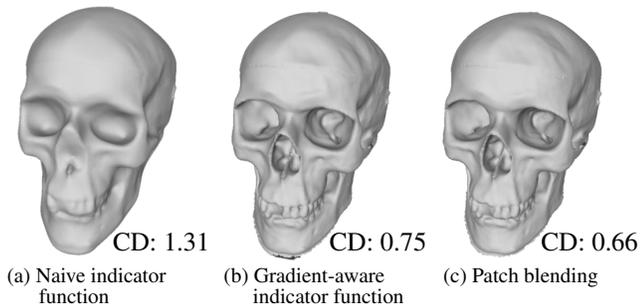

\begin{figure}
    \setlength{\abovecaptionskip}{1pt}
    \centering
    \begin{overpic}[width=\linewidth]{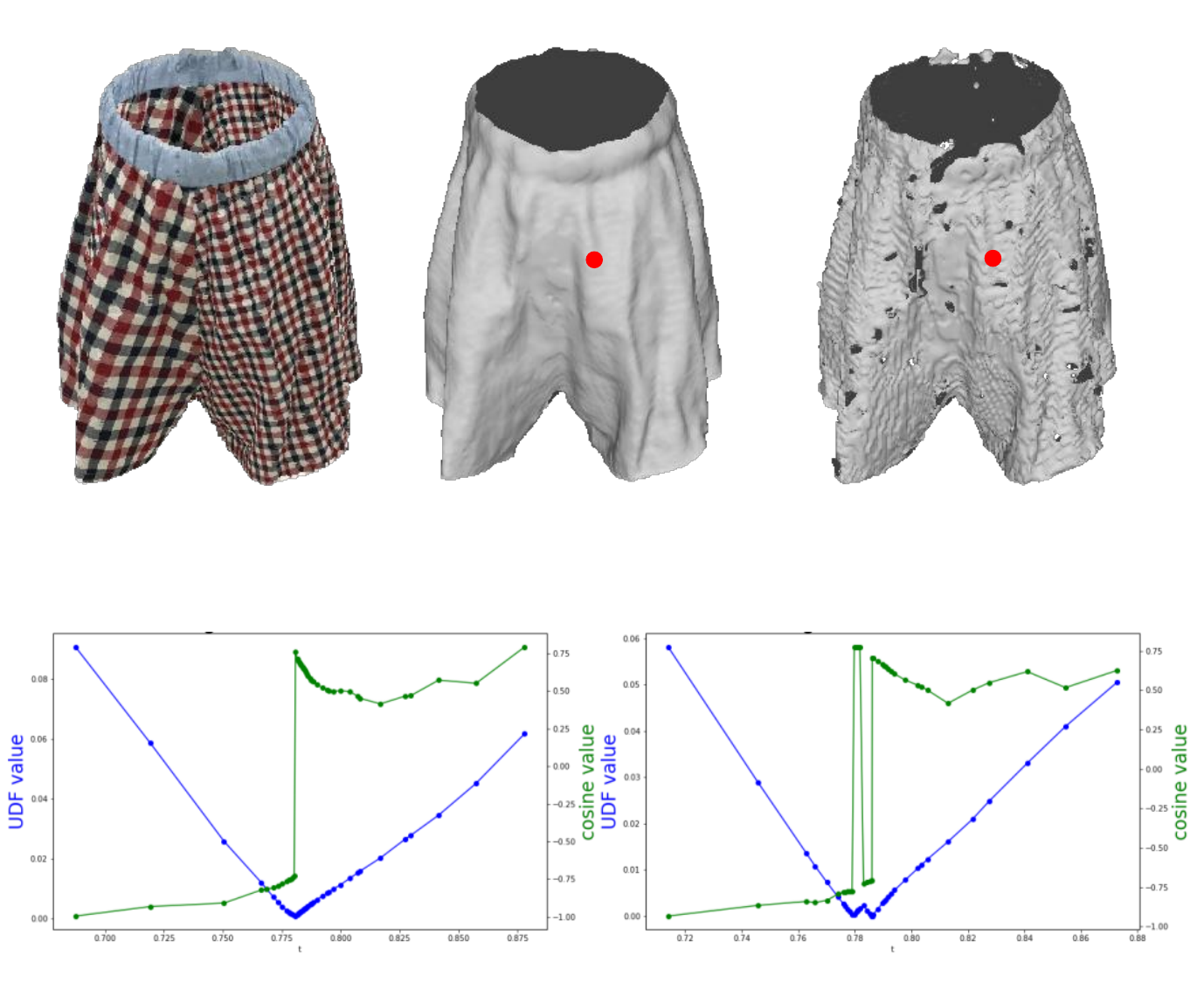}
    \put(95,93){CD: 2.47}  
    \put(175,93){CD: 2.49}
    \put(3,85){\small (a) Reference Image}  
    \put(85,85){\small (b) w/ regularizer}
    \put(160,85){\small (c) w/o regularizer}
    \put(15,73){\footnotesize  with iso-surface regularizer}
    \put(127,73){\footnotesize without iso-surface regularizer}
    \put(30,2){\small (d) The curves of UDF values and $\cos(\theta)$ values}
    \end{overpic}

    \caption{
    Ablation study on the iso-surface regularizer.
    }
    \label{fig:ablation_geo}
\end{figure}

\noindent
\textbf{Ablation studies.} We conduct ablation studies (Figure~\ref{fig:ablation} and Figure~\ref{fig:ablation_geo}) to investigate the individual contribution of the important designs of our method.
The first key design is the gradient-aware visibility indicator function for effectively indicating the surfaces. 
For convenience, we call the visibility indicator function that is not gradient-aware as naive indicator function.
As shown in Figure~\ref{fig:ablation}, the reconstruction result using the naive indicator function contains larger errors than the result using the gradient-aware indicator function, where the Chamfer Distance (CD) of the two reconstructions are 1.31 and 0.75 respectively (the lower is better).
We can also see that the reconstruction using the naive function lacks geometric details, where the teeth and eye holes of the Skull are not reconstructed accurately.
This is because the naive indicator function has inherent bias and the measured visibility is not consistent with the true surface intersection, thus leading to inaccurate reconstruction.

We leverage an iso-surface regularizer to make the UDF field more stable and accurate around zero level set. 
The reconstruction results with or without the regularizer are shown in Figure~\ref{fig:ablation_geo}.
Although the Chamfer Distances of the two results are close,
the mesh without the regularizer is much more noisy and has noticeable artifacts while the mesh with the regularizer is much cleaner and smoother.
To give an in-depth analysis, we plot the UDF values and the cosine values of the angle between the ray direction and UDF gradients along a selected ray (the marked red point of (b,c)).
We can see that, without the regularizer, the signs of the $\cos(\theta)$ values change multiple times around the zero level set, which indicates that the gradients are not stable. In contrast, with the regularizer, the gradients are more stable, where the gradient direction only flips at the surface intersection.

Also, inspired by ~\cite{long2022sparseneus}, the patch blending loss term is included in the optimization. With this term, the reconstruction can be further improved from 0.75 to 0.66 in terms of Chamfer Distance.

\begin{figure}
    \centering
    \begin{overpic}[width=0.95\linewidth]{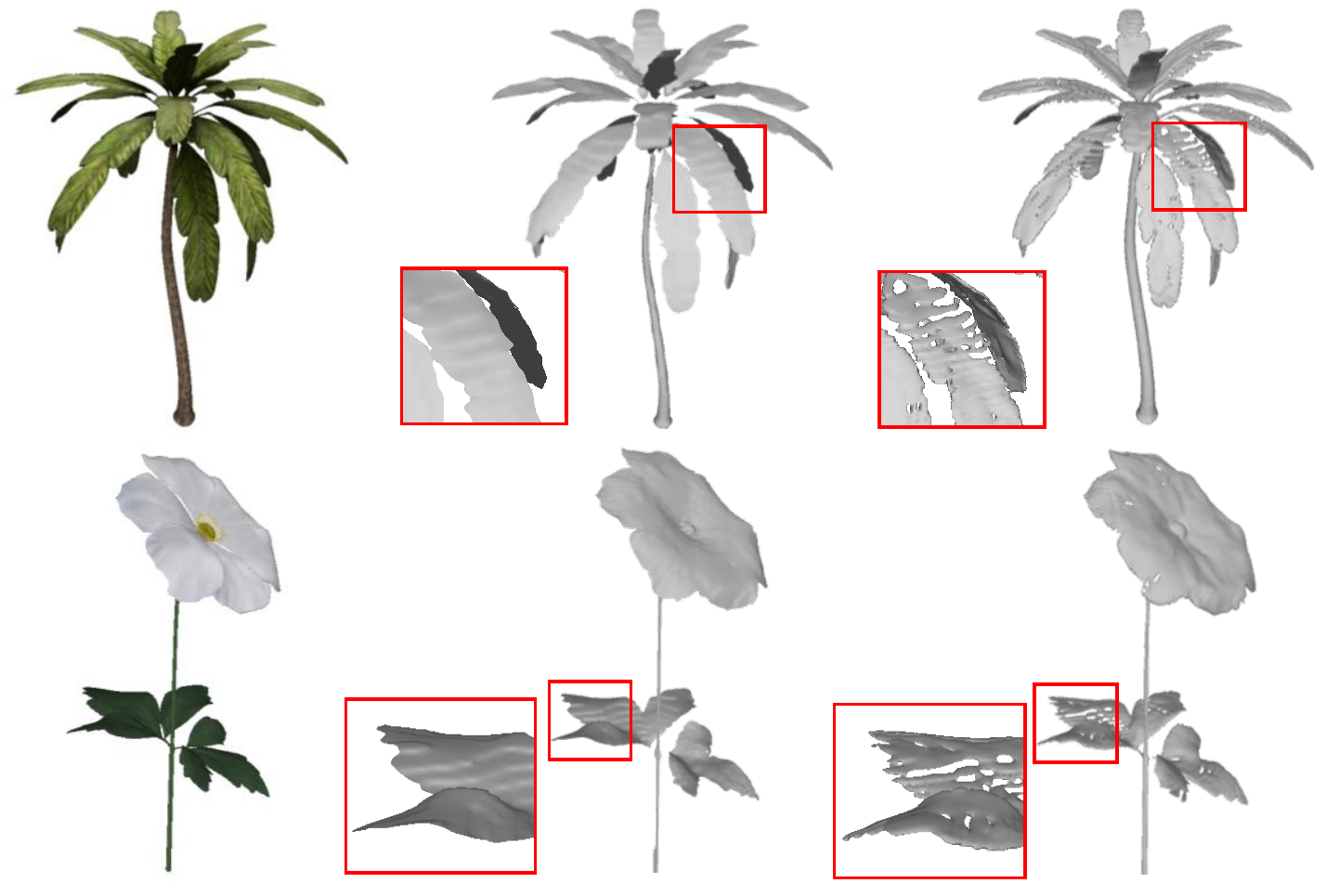}
    \put(0,-4){\small Reference Images}        
    \put(100,-4){\small Ours}
    \put(175,-4){\small NeuS~\cite{wang2021neus}}
    \end{overpic}

    \caption{
    Qualitative comparisons with NeuS on two plants.
    }
    \label{fig:plant_compare}
    \vspace{-5mm}
\end{figure}

\vspace{1mm}
\noindent
\textbf{More types of shapes.}
We additionally show results on several challenging objects with complex open surfaces.
Note {\em NeuralUDF} and NeuS~\cite{wang2021neus} adopt extra object masks in the optimization for the reconstruction.
As shown in Figure~\ref{fig:plant_compare}, our method successfully reconstructs the challenging plant models with non-manifold geometries, while the reconstruction results of NeuS~\cite{wang2021neus} are incomplete and contain numerous holes. 
This is because SDF is essentially not a suitable representation for many topologies like the thin leaves. Therefore, optimizing a SDF field for these kinds of models is deserved to be difficult and the reconstructed surfaces are naturally erroneous.

\noindent
\textbf{Limitation.}
Compared to the SDF based methods, the performance of {\em NeuralUDF} will degrade on textureless objects that lack enough distinguishable features.
{\em NeuralUDF} does not impose closed surface assumption and thus the representation has a higher degree of freedom. Without any topology priors, the surface optimization for textureless objects can be fairly difficult.


\section{Conclusions}
We introduce {\em NeuralUDF}, a NeRF-based method that uses neural UDF fields for mutliview reconstruction of surfaces with arbitrary topologies. Our key idea is to introduce a novel volume density function that effectively correlates the UDF representation with the volume rendering scheme to learn neural UDF fields. The experiments show that, unlike the SDF based works that can only model closed shapes, our method can reconstruct more types of shapes with either open or closed surfaces, thus extending the underlying geometric representation of neural volume rendering to various topologies.

\section*{Acknowlegements}
We thank Xiaoxu Meng for the help with the rendering code of DeepFashion3D dataset. 
Christian Theobalt was supported by ERC Consolidator Grant 770784. Lingjie Liu was supported by Lise Meitner Postdoctoral Fellowship.
Xiaoxiao Long was supported by the Hong Kong PhD Fellowship.

{\small
\bibliographystyle{ieee_fullname}

\begin{thebibliography}{10}\itemsep=-1pt

\bibitem{atzmon2020sal}
Matan Atzmon and Yaron Lipman.
\newblock Sal: Sign agnostic learning of shapes from raw data.
\newblock In {\em Proceedings of the IEEE/CVF Conference on Computer Vision and
  Pattern Recognition}, pages 2565--2574, 2020.

\bibitem{barron2021mip}
Jonathan~T Barron, Ben Mildenhall, Matthew Tancik, Peter Hedman, Ricardo
  Martin-Brualla, and Pratul~P Srinivasan.
\newblock Mip-nerf: A multiscale representation for anti-aliasing neural
  radiance fields.
\newblock In {\em Proceedings of the IEEE/CVF International Conference on
  Computer Vision}, pages 5855--5864, 2021.

\bibitem{bernardini1999ball}
Fausto Bernardini, Joshua Mittleman, Holly Rushmeier, Cl{\'a}udio Silva, and
  Gabriel Taubin.
\newblock The ball-pivoting algorithm for surface reconstruction.
\newblock {\em IEEE transactions on visualization and computer graphics},
  5(4):349--359, 1999.

\bibitem{campbell2008using}
Neill~DF Campbell, George Vogiatzis, Carlos Hern{\'a}ndez, and Roberto Cipolla.
\newblock Using multiple hypotheses to improve depth-maps for multi-view
  stereo.
\newblock In {\em European Conference on Computer Vision}, pages 766--779.
  Springer, 2008.

\bibitem{chen20223psdf}
Weikai Chen, Cheng Lin, Weiyang Li, and Bo Yang.
\newblock 3psdf: Three-pole signed distance function for learning surfaces with
  arbitrary topologies.
\newblock In {\em Proceedings of the IEEE/CVF Conference on Computer Vision and
  Pattern Recognition}, pages 18522--18531, 2022.

\bibitem{chen2019learning}
Zhiqin Chen and Hao Zhang.
\newblock Learning implicit fields for generative shape modeling.
\newblock In {\em Proceedings of the IEEE/CVF Conference on Computer Vision and
  Pattern Recognition}, pages 5939--5948, 2019.

\bibitem{chibane2020neural}
Julian Chibane, Gerard Pons-Moll, et~al.
\newblock Neural unsigned distance fields for implicit function learning.
\newblock {\em Advances in Neural Information Processing Systems},
  33:21638--21652, 2020.

\bibitem{darmon2021improving}
Fran{\c{c}}ois Darmon, B{\'e}n{\'e}dicte Bascle, Jean-Cl{\'e}ment Devaux,
  Pascal Monasse, and Mathieu Aubry.
\newblock Improving neural implicit surfaces geometry with patch warping.
\newblock {\em arXiv preprint arXiv:2112.09648}, 2021.

\bibitem{darmon2022improving}
Fran{\c{c}}ois Darmon, B{\'e}n{\'e}dicte Bascle, Jean-Cl{\'e}ment Devaux,
  Pascal Monasse, and Mathieu Aubry.
\newblock Improving neural implicit surfaces geometry with patch warping.
\newblock In {\em Proceedings of the IEEE/CVF Conference on Computer Vision and
  Pattern Recognition}, pages 6260--6269, 2022.

\bibitem{furukawa2009accurate}
Yasutaka Furukawa and Jean Ponce.
\newblock Accurate, dense, and robust multiview stereopsis.
\newblock {\em IEEE transactions on pattern analysis and machine intelligence},
  32(8):1362--1376, 2009.

\bibitem{galliani2015massively}
Silvano Galliani, Katrin Lasinger, and Konrad Schindler.
\newblock Massively parallel multiview stereopsis by surface normal diffusion.
\newblock In {\em Proceedings of the IEEE International Conference on Computer
  Vision}, pages 873--881, 2015.

\bibitem{gropp2020implicit}
Amos Gropp, Lior Yariv, Niv Haim, Matan Atzmon, and Yaron Lipman.
\newblock Implicit geometric regularization for learning shapes.
\newblock {\em arXiv preprint arXiv:2002.10099}, 2020.

\bibitem{gu2020cascade}
Xiaodong Gu, Zhiwen Fan, Siyu Zhu, Zuozhuo Dai, Feitong Tan, and Ping Tan.
\newblock Cascade cost volume for high-resolution multi-view stereo and stereo
  matching.
\newblock In {\em Proceedings of the IEEE/CVF Conference on Computer Vision and
  Pattern Recognition}, pages 2495--2504, 2020.

\bibitem{guillard2021meshudf}
Benoit Guillard, Federico Stella, and Pascal Fua.
\newblock Meshudf: Fast and differentiable meshing of unsigned distance field
  networks.
\newblock {\em arXiv preprint arXiv:2111.14549}, 2021.

\bibitem{jensen2014large}
Rasmus Jensen, Anders Dahl, George Vogiatzis, Engin Tola, and Henrik Aan{\ae}s.
\newblock Large scale multi-view stereopsis evaluation.
\newblock In {\em Proceedings of the IEEE conference on computer vision and
  pattern recognition}, pages 406--413, 2014.

\bibitem{ji2017surfacenet}
Mengqi Ji, Juergen Gall, Haitian Zheng, Yebin Liu, and Lu Fang.
\newblock Surfacenet: An end-to-end 3d neural network for multiview stereopsis.
\newblock In {\em Proceedings of the IEEE International Conference on Computer
  Vision}, pages 2307--2315, 2017.

\bibitem{ji2020surfacenet+}
Mengqi Ji, Jinzhi Zhang, Qionghai Dai, and Lu Fang.
\newblock Surfacenet+: An end-to-end 3d neural network for very sparse
  multi-view stereopsis.
\newblock {\em IEEE Transactions on Pattern Analysis and Machine Intelligence},
  43(11):4078--4093, 2020.

\bibitem{jiang2020sdfdiff}
Yue Jiang, Dantong Ji, Zhizhong Han, and Matthias Zwicker.
\newblock Sdfdiff: Differentiable rendering of signed distance fields for 3d
  shape optimization.
\newblock In {\em Proceedings of the IEEE/CVF Conference on Computer Vision and
  Pattern Recognition}, pages 1251--1261, 2020.

\bibitem{kar2017learning}
Abhishek Kar, Christian H{\"a}ne, and Jitendra Malik.
\newblock Learning a multi-view stereo machine.
\newblock {\em Advances in neural information processing systems}, 30, 2017.

\bibitem{kazhdan2006poisson}
Michael Kazhdan, Matthew Bolitho, and Hugues Hoppe.
\newblock Poisson surface reconstruction.
\newblock In {\em Proceedings of the fourth Eurographics symposium on Geometry
  processing}, volume~7, 2006.

\bibitem{kazhdan2013screened}
Michael Kazhdan and Hugues Hoppe.
\newblock Screened poisson surface reconstruction.
\newblock {\em ACM Transactions on Graphics (ToG)}, 32(3):1--13, 2013.

\bibitem{kellnhofer2021neural}
Petr Kellnhofer, Lars~C Jebe, Andrew Jones, Ryan Spicer, Kari Pulli, and Gordon
  Wetzstein.
\newblock Neural lumigraph rendering.
\newblock In {\em Proceedings of the IEEE/CVF Conference on Computer Vision and
  Pattern Recognition}, pages 4287--4297, 2021.

\bibitem{kutulakos2000theory}
Kiriakos~N Kutulakos and Steven~M Seitz.
\newblock A theory of shape by space carving.
\newblock {\em International journal of computer vision}, 38(3):199--218, 2000.

\bibitem{lee1980two}
Der-Tsai Lee and Bruce~J Schachter.
\newblock Two algorithms for constructing a delaunay triangulation.
\newblock {\em International Journal of Computer \& Information Sciences},
  9(3):219--242, 1980.

\bibitem{lhuillier2005quasi}
Maxime Lhuillier and Long Quan.
\newblock A quasi-dense approach to surface reconstruction from uncalibrated
  images.
\newblock {\em IEEE transactions on pattern analysis and machine intelligence},
  27(3):418--433, 2005.

\bibitem{lin2021barf}
Chen-Hsuan Lin, Wei-Chiu Ma, Antonio Torralba, and Simon Lucey.
\newblock Barf: Bundle-adjusting neural radiance fields.
\newblock In {\em Proceedings of the IEEE/CVF International Conference on
  Computer Vision}, pages 5741--5751, 2021.

\bibitem{liu2020neural}
Lingjie Liu, Jiatao Gu, Kyaw Zaw~Lin, Tat-Seng Chua, and Christian Theobalt.
\newblock Neural sparse voxel fields.
\newblock {\em Advances in Neural Information Processing Systems},
  33:15651--15663, 2020.

\bibitem{liu2020dist}
Shaohui Liu, Yinda Zhang, Songyou Peng, Boxin Shi, Marc Pollefeys, and Zhaopeng
  Cui.
\newblock Dist: Rendering deep implicit signed distance function with
  differentiable sphere tracing.
\newblock In {\em Proceedings of the IEEE/CVF Conference on Computer Vision and
  Pattern Recognition}, pages 2019--2028, 2020.

\bibitem{long2021adaptive}
Xiaoxiao Long, Cheng Lin, Lingjie Liu, Wei Li, Christian Theobalt, Ruigang
  Yang, and Wenping Wang.
\newblock Adaptive surface normal constraint for depth estimation.
\newblock In {\em Proceedings of the IEEE/CVF International Conference on
  Computer Vision}, pages 12849--12858, 2021.

\bibitem{long2022sparseneus}
Xiaoxiao Long, Cheng Lin, Peng Wang, Taku Komura, and Wenping Wang.
\newblock Sparseneus: Fast generalizable neural surface reconstruction from
  sparse views.
\newblock {\em arXiv preprint arXiv:2206.05737}, 2022.

\bibitem{long2021multi}
Xiaoxiao Long, Lingjie Liu, Wei Li, Christian Theobalt, and Wenping Wang.
\newblock Multi-view depth estimation using epipolar spatio-temporal networks.
\newblock In {\em Proceedings of the IEEE/CVF Conference on Computer Vision and
  Pattern Recognition}, pages 8258--8267, 2021.

\bibitem{long2020occlusion}
Xiaoxiao Long, Lingjie Liu, Christian Theobalt, and Wenping Wang.
\newblock Occlusion-aware depth estimation with adaptive normal constraints.
\newblock In {\em European Conference on Computer Vision}, pages 640--657.
  Springer, 2020.

\bibitem{mescheder2019occupancy}
Lars Mescheder, Michael Oechsle, Michael Niemeyer, Sebastian Nowozin, and
  Andreas Geiger.
\newblock Occupancy networks: Learning 3d reconstruction in function space.
\newblock In {\em Proceedings of the IEEE/CVF Conference on Computer Vision and
  Pattern Recognition}, pages 4460--4470, 2019.

\bibitem{michalkiewicz2019implicit}
Mateusz Michalkiewicz, Jhony~K Pontes, Dominic Jack, Mahsa Baktashmotlagh, and
  Anders Eriksson.
\newblock Implicit surface representations as layers in neural networks.
\newblock In {\em Proceedings of the IEEE/CVF International Conference on
  Computer Vision}, pages 4743--4752, 2019.

\bibitem{mildenhall2020nerf}
Ben Mildenhall, Pratul~P Srinivasan, Matthew Tancik, Jonathan~T Barron, Ravi
  Ramamoorthi, and Ren Ng.
\newblock Nerf: Representing scenes as neural radiance fields for view
  synthesis.
\newblock In {\em European conference on computer vision}, pages 405--421.
  Springer, 2020.

\bibitem{niemeyer2021giraffe}
Michael Niemeyer and Andreas Geiger.
\newblock Giraffe: Representing scenes as compositional generative neural
  feature fields.
\newblock In {\em Proceedings of the IEEE/CVF Conference on Computer Vision and
  Pattern Recognition}, pages 11453--11464, 2021.

\bibitem{niemeyer2020differentiable}
Michael Niemeyer, Lars Mescheder, Michael Oechsle, and Andreas Geiger.
\newblock Differentiable volumetric rendering: Learning implicit 3d
  representations without 3d supervision.
\newblock In {\em Proceedings of the IEEE/CVF Conference on Computer Vision and
  Pattern Recognition}, pages 3504--3515, 2020.

\bibitem{oechsle2021unisurf}
Michael Oechsle, Songyou Peng, and Andreas Geiger.
\newblock Unisurf: Unifying neural implicit surfaces and radiance fields for
  multi-view reconstruction.
\newblock In {\em Proceedings of the IEEE/CVF International Conference on
  Computer Vision}, pages 5589--5599, 2021.

\bibitem{park2019deepsdf}
Jeong~Joon Park, Peter Florence, Julian Straub, Richard Newcombe, and Steven
  Lovegrove.
\newblock Deepsdf: Learning continuous signed distance functions for shape
  representation.
\newblock In {\em Proceedings of the IEEE/CVF Conference on Computer Vision and
  Pattern Recognition}, pages 165--174, 2019.

\bibitem{park2021nerfies}
Keunhong Park, Utkarsh Sinha, Jonathan~T Barron, Sofien Bouaziz, Dan~B Goldman,
  Steven~M Seitz, and Ricardo Martin-Brualla.
\newblock Nerfies: Deformable neural radiance fields.
\newblock In {\em Proceedings of the IEEE/CVF International Conference on
  Computer Vision}, pages 5865--5874, 2021.

\bibitem{schonberger2016structure}
Johannes~L Schonberger and Jan-Michael Frahm.
\newblock Structure-from-motion revisited.
\newblock In {\em Proceedings of the IEEE conference on computer vision and
  pattern recognition}, pages 4104--4113, 2016.

\bibitem{schonberger2016pixelwise}
Johannes~L Sch{\"o}nberger, Enliang Zheng, Jan-Michael Frahm, and Marc
  Pollefeys.
\newblock Pixelwise view selection for unstructured multi-view stereo.
\newblock In {\em European Conference on Computer Vision}, pages 501--518.
  Springer, 2016.

\bibitem{seitz1999photorealistic}
Steven~M Seitz and Charles~R Dyer.
\newblock Photorealistic scene reconstruction by voxel coloring.
\newblock {\em International Journal of Computer Vision}, 35(2):151--173, 1999.

\bibitem{sun2021neuralrecon}
Jiaming Sun, Yiming Xie, Linghao Chen, Xiaowei Zhou, and Hujun Bao.
\newblock Neuralrecon: Real-time coherent 3d reconstruction from monocular
  video.
\newblock In {\em Proceedings of the IEEE/CVF Conference on Computer Vision and
  Pattern Recognition}, pages 15598--15607, 2021.

\bibitem{tola2012efficient}
Engin Tola, Christoph Strecha, and Pascal Fua.
\newblock Efficient large-scale multi-view stereo for ultra high-resolution
  image sets.
\newblock {\em Machine Vision and Applications}, 23(5):903--920, 2012.

\bibitem{wang2021neus}
Peng Wang, Lingjie Liu, Yuan Liu, Christian Theobalt, Taku Komura, and Wenping
  Wang.
\newblock Neus: Learning neural implicit surfaces by volume rendering for
  multi-view reconstruction.
\newblock {\em Advances in Neural Information Processing Systems}, 34, 2021.

\bibitem{yao2018mvsnet}
Yao Yao, Zixin Luo, Shiwei Li, Tian Fang, and Long Quan.
\newblock Mvsnet: Depth inference for unstructured multi-view stereo.
\newblock In {\em Proceedings of the European Conference on Computer Vision
  (ECCV)}, pages 767--783, 2018.

\bibitem{yao2019recurrent}
Yao Yao, Zixin Luo, Shiwei Li, Tianwei Shen, Tian Fang, and Long Quan.
\newblock Recurrent mvsnet for high-resolution multi-view stereo depth
  inference.
\newblock In {\em Proceedings of the IEEE/CVF Conference on Computer Vision and
  Pattern Recognition}, pages 5525--5534, 2019.

\bibitem{yariv2021volume}
Lior Yariv, Jiatao Gu, Yoni Kasten, and Yaron Lipman.
\newblock Volume rendering of neural implicit surfaces.
\newblock {\em Advances in Neural Information Processing Systems}, 34, 2021.

\bibitem{yariv2020multiview}
Lior Yariv, Yoni Kasten, Dror Moran, Meirav Galun, Matan Atzmon, Basri Ronen,
  and Yaron Lipman.
\newblock Multiview neural surface reconstruction by disentangling geometry and
  appearance.
\newblock {\em Advances in Neural Information Processing Systems},
  33:2492--2502, 2020.

\bibitem{zhang2021learning}
Jingyang Zhang, Yao Yao, and Long Quan.
\newblock Learning signed distance field for multi-view surface reconstruction.
\newblock In {\em Proceedings of the IEEE/CVF International Conference on
  Computer Vision}, pages 6525--6534, 2021.

\bibitem{zhang2020nerf++}
Kai Zhang, Gernot Riegler, Noah Snavely, and Vladlen Koltun.
\newblock Nerf++: Analyzing and improving neural radiance fields.
\newblock {\em arXiv preprint arXiv:2010.07492}, 2020.

\bibitem{zhu2020deep}
Heming Zhu, Yu Cao, Hang Jin, Weikai Chen, Dong Du, Zhangye Wang, Shuguang Cui,
  and Xiaoguang Han.
\newblock Deep fashion3d: A dataset and benchmark for 3d garment reconstruction
  from single images.
\newblock In {\em European Conference on Computer Vision}, pages 512--530.
  Springer, 2020.

\end{thebibliography}

}

\newpage

\section*{Supplementary Material}

\subsection*{Loss Functions}
We train {\em NeuralUDF} by enforcing the consistency of the rendered colors and the ground truth colors of the input images without using 3D ground truth shapes. 
Specifically, we optimize our neural networks and two trainable probabilistic parameters $\kappa$  and $\beta$ by randomly sampling a batch of pixels and their corresponding rays.
Suppose that the size of sampling points in a ray is $N$ and the batch size is $M$.

The overall loss  function is defined as a weighted sum of several loss terms:
\begin{equation}
\label{total_loss}
    \mathcal{L}=\mathcal{L}_{\text {color }} + \lambda_{0} \mathcal{L}_{\text {patch }} +\lambda_{1} \mathcal{L}_{\text {eik }} +\lambda_{2} \mathcal{L}_{\text {reg}} \left(+  \gamma \mathcal{L}_{\text {mask }} \right).
\end{equation}

The color loss $\mathcal{L}_{\text{color}}$ is defined as
\begin{equation}
\mathcal{L}_{\text {color }}=\frac{1}{M} \sum_k |\hat{C}_k- C_k|,
\end{equation}
where $\hat{C}_k$ is the rendered color of the $k$-th pixel in a batch, and $C_k$ is the corresponding ground truth color.

Same as the prior works~\citeSupp{yariv2020multiview,wang2021neus,yariv2021volume}, we utilize the Eikonal term on the sampled points to regularize the UDF field $f_u$ to have unit norm of gradients:
\begin{equation}
\mathcal{L}_{eik}=\frac{1}{N M} \sum_{k, i}\left(\left\|\nabla f_u\left( \mathbf{r}_{k}(t_{k,i}) \right)\right\|_2-1\right)^2.
\end{equation}

Optionally, if extra object masks are provided, we can adopt a mask loss:
\begin{equation}
\mathcal{L}_{\text {mask }}=\operatorname{BCE}\left(M_k, \hat{O}_k\right),
\end{equation}
where $M_k$ is the mask value of the k-th pixel, $\hat{O}_k=\sum_{i=1}^n T_{k, i} \sigma_{k, i} \delta_{k, i}$ is the sum of weights of volume rendering along the camera ray, and $\operatorname{BCE}$ is the binary cross entropy loss.

Besides minimizing the differences between the rendered color and ground truth color pixel by pixel, we render the colors of a local patch using the patch blending strategy used in ~\citeSupp{long2022sparseneus} and enforce the consistency of the rendered colors and ground truth colors at the patch level by including the loss term:
\begin{equation}
\mathcal{L}_{\text {patch}}=\frac{1}{M} \sum_k \mathcal{R}\left(\hat{P}_k, P_k\right),
\end{equation}
where the $\hat{P}_k$ are the rendered colors of a patch centering the $k$-th sampled pixel and the $P_k$ are the corresponding ground truth colors. Empirically, we choose $\mathcal{R}$ to be the Structural Similarity Index Measure (SSIM) metric.

\subsection*{Zero Level Set Analysis}
\begin{figure}
    \centering
    \begin{overpic}[width=0.5\linewidth]{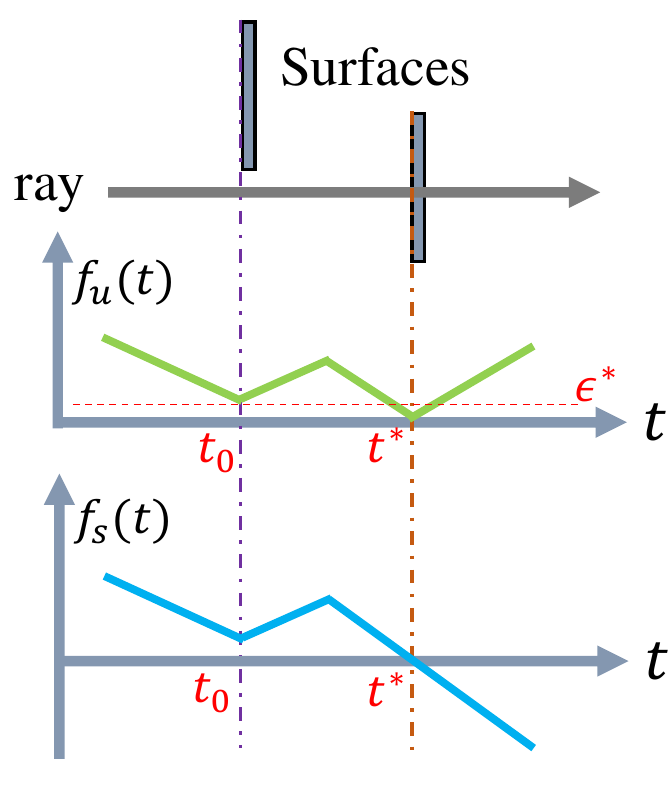}
    
    \end{overpic}
    \caption{
    Illustration about zero level sets of UDF and SDF. }
    \label{fig:zero_level_set}
    \vspace{-4mm}
\end{figure}
The key point of applying volume rendering to UDF is finding the intersection point $t^{*}$ of the ray and the surface, that is, the zero level set of UDF.
However, it's challenging for UDF to find the intersection point $t^{*}$, since it's impossible to locate a point with exact zero UDF value.
A common practice is to find points whose UDF values are smaller than a pre-defined threshold $\epsilon$. As a result, the approximated surfaces will be relaxed from exact zero level set to be a $\epsilon$-bounded surface band $\mathcal{S}=\left\{\mathbf{x} \in \mathbb{R}^3 \mid f_{u}(\mathbf{x})< \epsilon \right\}$.

However, finding an appropriate $\epsilon$ is difficult for the reconstruction task. 
A typical case is shown in Figure~\ref{fig:zero_level_set}, a ray first passes by a surface at point $t_{0}$ and then hit on a surface at point $t^{*}$.
It's easy for SDF to locate the intersection point $t^{*}$ by finding the sign flipping of SDF values.
On the contrary, UDF gives all positive values, thus leading to the difficulty in distinguishing the two points $t_{0}$ and $t^{*}$.
To locate the true surface intersection point, the pre-defined $\epsilon$ should be smaller than the udf value $f_u(\mathbf{r}(t_0))$, otherwise the point $t_0$ will be mistakenly located as the intersection point, causing inaccurate volume rendering.
Moreover, for objects with complex geometries, a fixed $\epsilon$ cannot be adapted to different optimization stages.
To tackle these problems, we leverage probabilistic models to design a differentiable indicator function $\Psi(t)$ to locate the surface intersection points.

\subsection*{More Implementation Details}
\paragraph{Network architecture}
Following the prior works~\citeSupp{wang2021neus,yariv2021volume,yariv2020multiview}, we use two MLPs to encode UDF and color respectively. The unsigned distance function $f_u$ is modeled by an MLP that consists of 8 hidden layers with hidden size of 256. A skip connection~\citeSupp{park2019deepsdf} is used to connect the output of the fourth layer and the input. 
The color function $c$ is modeled by an MLP with 8 hidden layers with size of 256, which takes the combination of spatial location $\mathbf{r}(t)$, the view direction $\mathbf{v}$, and a 256-dimension feature vector from the UDF MLP as input.
Same as the prior works~\citeSupp{wang2021neus,yariv2021volume,yariv2020multiview}, we apply positional encoding to the spatial location $\mathbf{r}(t)$ with 6 frequencies and to the view direction $\mathbf{v}$ with 4 frequencies.

\paragraph{Training details}
The Adam optimizer is used to train our networks. We set the learning rate of the UDF MLP as 1e-4 and the other networks and trainable parameters as 5e-4, and the learning rates are controlled by the cosine decay schedule.
We set the constant parameter $\alpha$ of the Eq.~\ref{existence_probability} as 20, and the constant parameter $\tau$ of the Eq.~\ref{iso_regularizer} as 25000.
The loss weights of Eq.~\ref{total_loss} are set as $\lambda_0=0.1, \lambda_1=0.1, \lambda_2 = 0.0$ for the DTU~\citeSupp{jensen2014large} dataset, and set as $\lambda_0=0.0, \lambda_1=0.01, \lambda_2 = 0.01$ for the DeepFashion3D~\citeSupp{zhu2020deep} dataset. The $\lambda_2$ sometimes should be tuned for some cases to achieve better reconstruction.
$\gamma$ will be set to 0.1 if mask loss term is adopted.

\paragraph{Hierarchical sampling}
To sample more points close to the surfaces, we adopt a hierarchical sampling strategy. We first uniformly sample 64 points along the ray, then we iteratively conduct importance sampling for 5 times.
The coarse probability estimation in the $i$-th iteration is computed by density function with fixed $\kappa$ and $\beta$, which are set as $32 \times 2^{i}$ and $32 \times 2^{i+1}$ respectively. We sample extra 16 points in each iteration, and the total number of sampled points is 144. 
We scale the scene to be reconstructed in the unit sphere, and model the space outside the unit sphere using NeRF++~\citeSupp{zhang2020nerf++}.

\subsection*{More Results}
\begin{figure*}[tp!]
    \centering
    \begin{overpic}[width=0.95\linewidth]{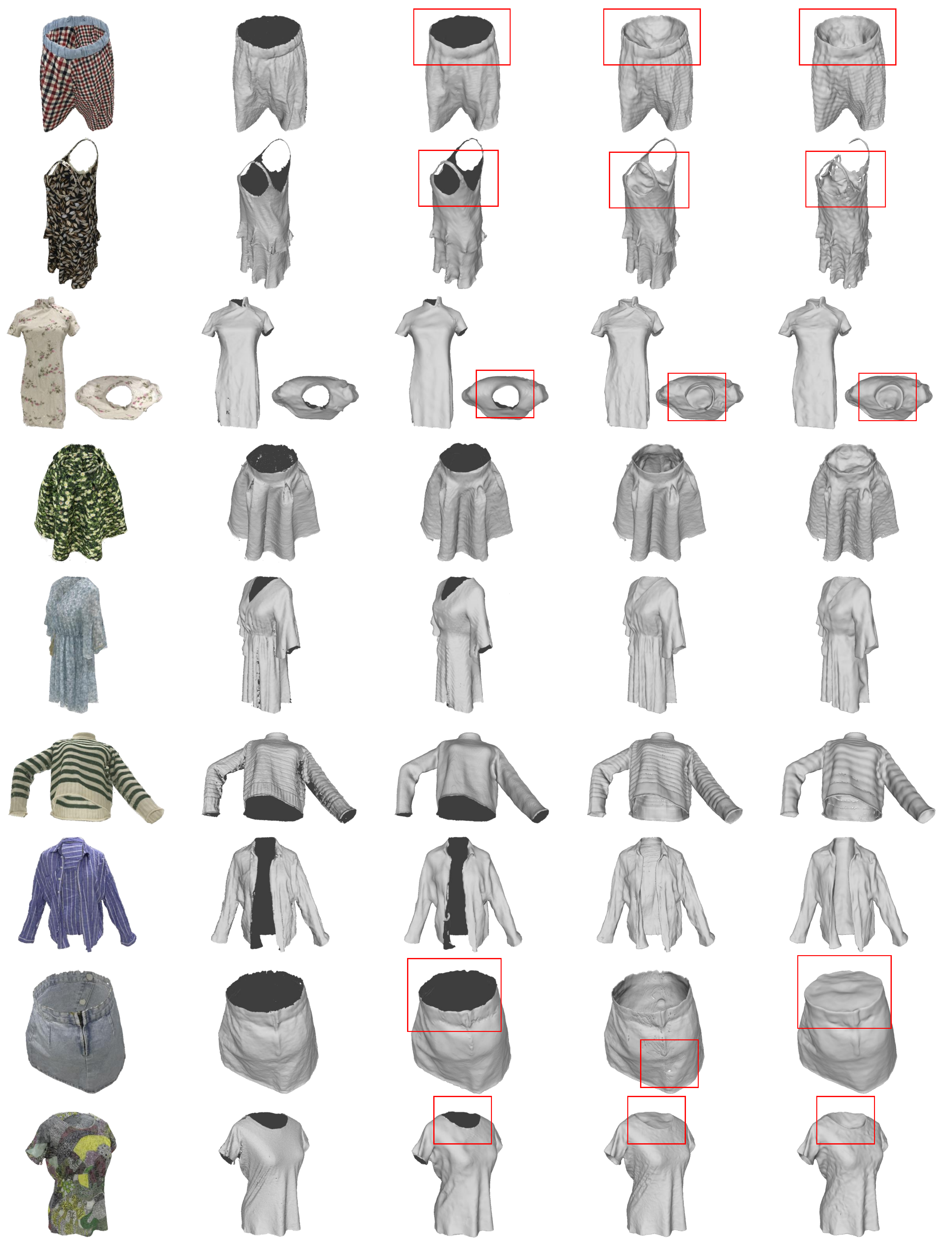}
    \put(-8,35){\small R9} 
    \put(-8,110){\small R8} 
    \put(-8,180){\small R7} 
    \put(-8,240){\small R6} 
    \put(-8,300){\small R5} 
    \put(-8,370){\small R4} 
    \put(-8,440){\small R3} 
    \put(-8,520){\small R2} 
    \put(-8,580){\small R1}

    \put(5,-2){\small Reference Images}  
    \put(115,-2){\small GT meshes}
    \put(220,-2){\small Ours}
    \put(310,-2){\small NeuS~\cite{wang2021neus}}
    \put(405,-2){\small VolSDF~\cite{yariv2021volume}}
    
    \put(5, 625){\small Reference Images}  
    \put(115,625){\small GT meshes}
    \put(220,625){\small Ours}
    \put(310,625){\small NeuS~\cite{wang2021neus}}
    \put(405,625){\small VolSDF~\cite{yariv2021volume}}
    
    \end{overpic}
    \caption{
    Qualitative comparisons with SOTAs on DeepFashion3D~\cite{zhu2020deep} dataset. The GT meshes are reconstructed from GT point clouds provided by DeepFashion3D dataset using Ball Pivoting algorithm~\cite{bernardini1999ball}.}
    \label{fig:supp_garment_compare}
\end{figure*}

Here we show the remaining qualitative comparisons of DeepFashion3D~\citeSupp{zhu2020deep} and DTU~\citeSupp{jensen2014large} datasets.
Since the DeepFashion3D dataset only provides ground truth point clouds, we use Ball Pivoting meshing algorithm~\citeSupp{bernardini1999ball} to reconstruct meshes for better visualization.
As shown in Figure~\ref{fig:supp_garment_compare}, the SDF based methods model all the shapes as closed surfaces and cannot faithfully recover the captured objects, thus causing erroneous geometries. 
This is because these methods enforce strong assumption on the closeness of the target shapes.
For example, for the first row and the second row, NeuS and VolSDF give closed surfaces for the dress and the trouser where their opening boundaries are even incorrectly connected, thus leading to noticeable errors.
For the 8-th row, NeuS attempts to use a double-layered surface to represent the thin cloth; optimizing the tiny SDF band is difficult and its result contains holes.
In contrast, our method faithfully reconstruct the geometries with clean and sharp open boundaries.

\begin{figure*}
    \centering
    \begin{overpic}[width=0.95\linewidth]{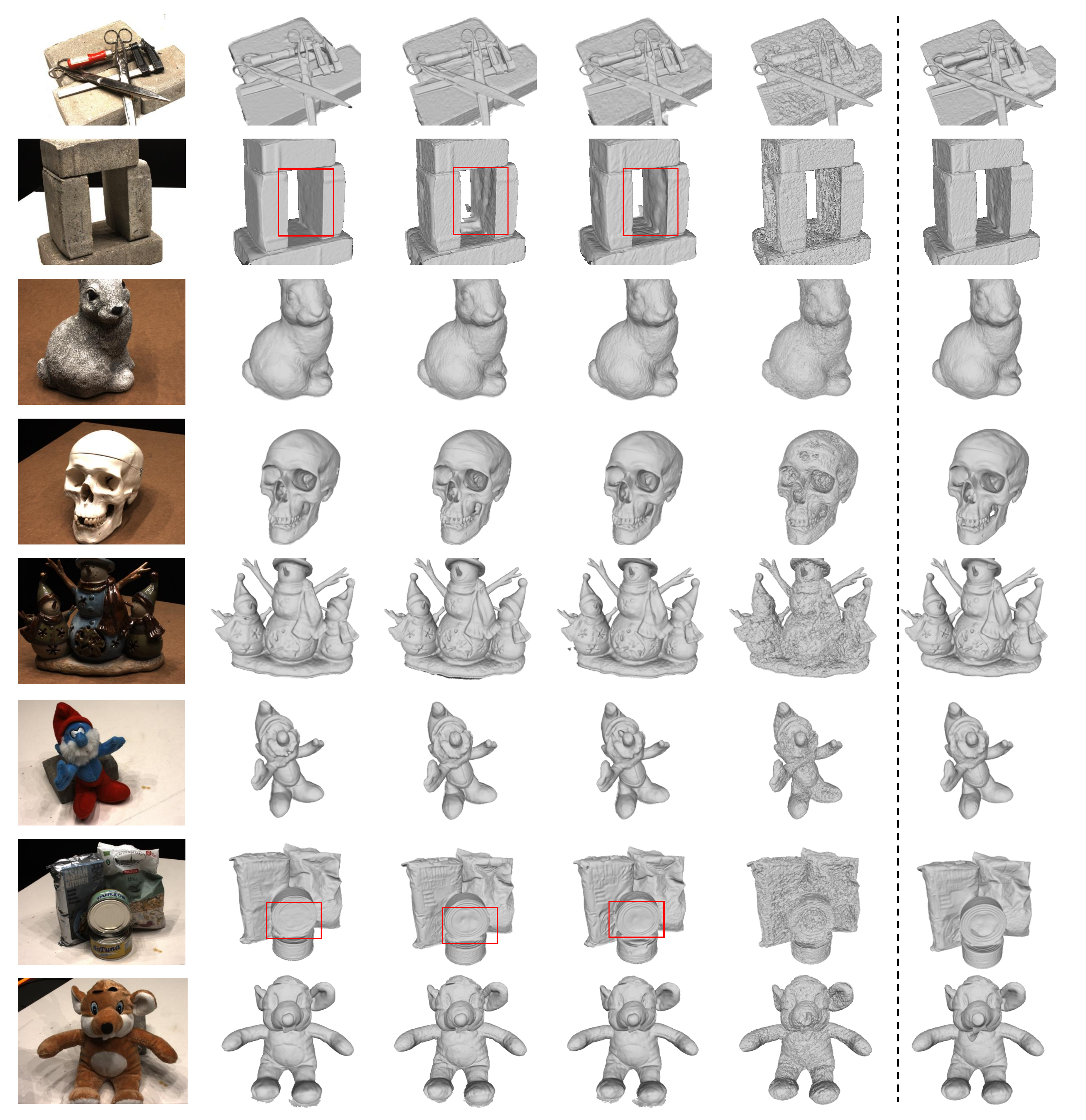}
    
    \put(-5,30){\small R8} 
    \put(-5,90){\small R7} 
    \put(-5,150){\small R6} 
    \put(-5,215){\small R5} 
    \put(-5,270){\small R4} 
    \put(-5,335){\small R3} 
    \put(-5,395){\small R2} 
    \put(-5,455){\small R1}

    \put(15,485){\small Reference Images}  
    \put(120,485){\small Ours}
    \put(185,485){\small NeuS~\cite{wang2021neus}}
    \put(257,485){\small VolSDF~\cite{yariv2021volume}}
    \put(335,485){\small NeRF~\cite{mildenhall2020nerf}}
    \put(412,485){\small IDR~\cite{yariv2020multiview}}

    \put(15,-4){\small Reference Images}  
    \put(120,-4){\small Ours}
    \put(185,-4){\small NeuS~\cite{wang2021neus}}
    \put(257,-4){\small VolSDF~\cite{yariv2021volume}}
    \put(335,-4){\small NeRF~\cite{mildenhall2020nerf}}
    \put(412,-4){\small IDR~\cite{yariv2020multiview}}
    \end{overpic}
    \caption{
    Qualitative comparisons with SOTAs on DTU~\cite{jensen2014large} dataset (Part 1/2).
    }
    \label{fig:supp_dtu_compare_1}
\end{figure*}

\begin{figure*}
    \centering
    \begin{overpic}[width=0.95\linewidth]{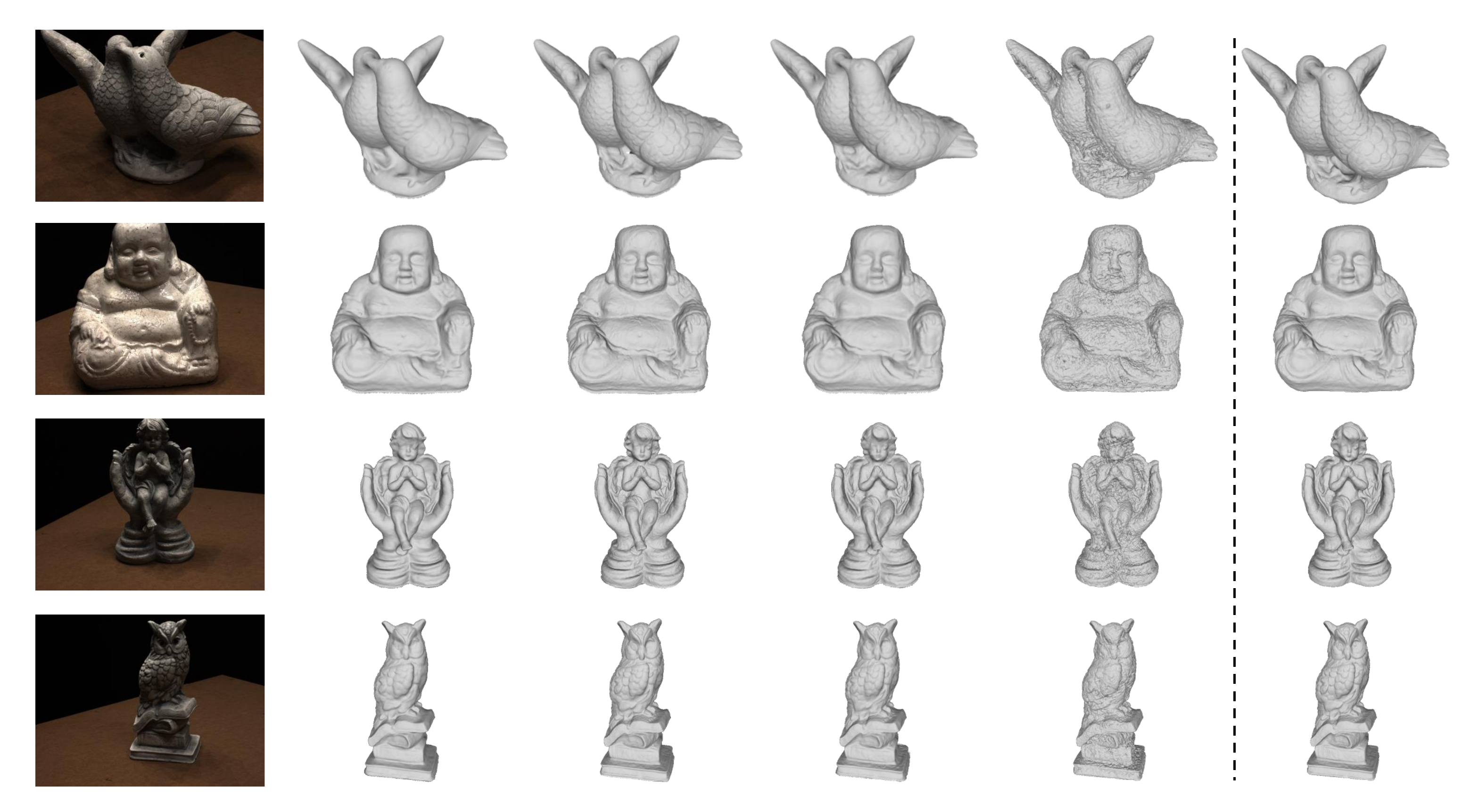}
    \put(15,-4){\small Reference Images}  
    \put(120,-4){\small Ours}
    \put(185,-4){\small NeuS~\cite{wang2021neus}}
    \put(257,-4){\small VolSDF~\cite{yariv2021volume}}
    \put(335,-4){\small NeRF~\cite{mildenhall2020nerf}}
    \put(412,-4){\small IDR~\cite{yariv2020multiview}}
    \end{overpic}
    \caption{
    Qualitative comparisons with SOTAs on DTU~\cite{jensen2014large} dataset (Part 2/2).
    }
    \label{fig:supp_dtu_compare_2}
\end{figure*}

The remaining comparisons of DTU~\citeSupp{jensen2014large} dataset are plotted in Figure~\ref{fig:supp_dtu_compare_1} and Figure~\ref{fig:supp_dtu_compare_2}.
Unlike the SDF based methods that are particularly designed for closed surfaces, our method adopts UDF as surface representation and does not enforce the closed surface assumption. 
Although our method doesn't impose the closed surface constraint, our method still achieves comparable performance with the SDF based methods.
As shown in the Figure~\ref{fig:supp_dtu_compare_1}, for the second row, NeuS and VolSDF fails to accurately reconstruct the challenging side part of the Brick (the red marked box) since this part is occluded in many input images and lacks enough photography consistency, while our result is more accurate and contain less artifacts.
For the 7-th row, the Can with reflecting material, the reconstruction results of NeuS and VolSDF have indentation artifacts on the surface while our method successfully reconstructs the flat and smooth surface.

The experiments validate that, unlike that the SDF based methods are limited to closed shapes, our method is more flexible and can cope with more types of shapes with either open or close surfaces, thus extending the underlying geometric representation of neural volume rendering to various topologies.



\end{document}